\documentclass[preprint, sts]{imsart}

\usepackage{amsthm,amsmath,natbib}
\RequirePackage[colorlinks,citecolor=blue,urlcolor=blue]{hyperref}
\usepackage{enumitem}

\usepackage{enumitem}
\usepackage{graphicx}
\usepackage{float}
\usepackage{amssymb}
\usepackage{macros}
\usepackage{subfigure}
\usepackage{stackrel}
\usepackage{url}

\startlocaldefs
\newcommand{\sgn}{\operatorname{sgn}}

\newtheorem{thm}{Theorem}[section]

\newtheorem{theorem}{Theorem}[section]

\newtheorem{lemma}[theorem]{Lemma}

\endlocaldefs

\begin{document}

\begin{frontmatter}

\title{On the Sensitivity of the Lasso to the Number of Predictor Variables}
\runtitle{The Sensitivity of the Lasso}

\author{\fnms{Cheryl J.} \snm{Flynn}\corref{}\ead[label=e1]{cflynn@stern.nyu.edu}},
\author{\fnms{Clifford M.} \snm{Hurvich}\ead[label=e2]{churvich@stern.nyu.edu}}
\and
\author{\fnms{Jeffrey S.} \snm{Simonoff}\ead[label=e3]{jsimonof@stern.nyu.edu}}

\address{\printead{e1}.}
\affiliation{New York University}

\runauthor{C. J. Flynn, C. M. Hurvich and J. S. Simonoff}

\begin{abstract}
The Lasso is a computationally efficient regression regularization procedure that can produce sparse estimators when the number of predictors $(p)$ is large.
Oracle inequalities provide probability loss bounds for the Lasso estimator at a deterministic choice of the regularization parameter. These bounds tend to zero if $p$ is appropriately controlled, and are thus commonly cited as theoretical justification for the Lasso and its ability to handle high-dimensional settings.  Unfortunately, in practice the regularization parameter is not selected to be a deterministic quantity, but is instead chosen using a random, data-dependent procedure.  To address this shortcoming of previous theoretical work, we study the loss of the Lasso estimator when tuned optimally for prediction.  Assuming orthonormal predictors and a sparse true model, we prove that the probability that the best possible predictive performance of the Lasso deteriorates as $p$ increases is positive and can be arbitrarily close to one given a sufficiently high signal to noise ratio and sufficiently large $p$.  We further demonstrate empirically that the amount of deterioration in performance can be far worse than the oracle inequalities suggest and provide a real data example where deterioration is observed.
\end{abstract}

\begin{keyword}
\kwd{Least Absolute Shrinkage and Selection Operator (Lasso)}
\kwd{Oracle Inequalities}
\kwd{High-Dimensional Data}
\end{keyword}

\end{frontmatter}

\section{Introduction}\label{intro}
Regularization methods perform model selection subject to the choice of a regularization parameter, and are commonly used when the number of predictor variables is too large to consider all subsets.  In regularized regression, these methods operate by minimizing the penalized least squares function
\begin{equation}\label{e:lasso}
\frac{1}{2}||\vy - \mX \vbeta||^2 + \lambda Pen(\vbeta)
\end{equation}
where $\vy$ is a $n \times 1$ response vector, $\mX$ is a $n \times p$ deterministic matrix of predictor variables, $\vbeta$ is a $p \times 1$ vector of coefficients, and $Pen(\cdot)$ is a penalty function.  A common choice for the penalty function is the $l_1$ norm of the coefficients.  This penalty function was proposed by \citet{tibshirani96} and termed the Lasso (Least absolute shrinkage and selection operator).  The solution to the Lasso is sparse in that it automatically sets some of the estimated coefficients equal to zero, and the entire regularization path can be found using the computationally efficient Lars algorithm \citep{efron04}.  Given its computational advantages, understanding the theoretical properties of the Lasso is an important area of research.

This paper focuses on the predictive performance of the Lasso and the impact of regularization.  To that end, we evaluate the Lasso-estimated models using the $l_2$-loss function.  We assume that the true data generating process is
\begin{equation}\label{e:dgp}
\vy = \vmu+\mathbf{\varepsilon}
\end{equation}
where $\vmu$ is a $n \times 1$ unknown mean vector and $\mathbf{\varepsilon}$ is a $n \times 1$ random noise vector.
Then the $l_2$-loss
is defined as
\begin{equation}\label{e:loss}
L_p(\lambda) = \frac{ || \vmu - \hat \vmu_\lambda ||^2 }{n} =
    \frac{ || \vmu - \mX \hat \vbeta_\lambda ||^2 }{n}
\end{equation}
where $\hat \vbeta_\lambda$ is the Lasso estimated vector of coefficients for a specific choice of the regularization parameter $\lambda \in [0,\infty)$ and $||\cdot||^2$ is the squared Euclidean norm.  Here we subscript the loss by $p$ to emphasize that the loss at a particular value of $\lambda$ depends on the number of predictor variables.  If the true model is included among the candidate models, then $\vmu = \mX \vbeta_0$ for some unknown true coefficient vector $\vbeta_0$ and the $l_2$-loss function takes the form
\[
L_p(\lambda) =
    \frac{ || \mX(\vbeta_0 - \hat \vbeta_\lambda)||^2 }{n}.
\]
To be consistent with most modern applications, we allow $\vbeta_0$ to be sparse and assume that it has $p_0 \leq p$ non-zero entries.

Probability loss bounds exist for the Lasso in this setting (e.g., \citealp{candes09}, \citealp{bickel10}, and \citealp{buhlmann11}).  Roughly, for a particular deterministic choice, $\lambda^0$, of $\lambda$,  these probability bounds are of the form
\begin{equation}\label{e:loss_bound_rough}
L_p(\lambda^0) \leq k \sigma^2 \frac{\log(p) p_0}{n}
\end{equation}
(p. 102, \citealp{buhlmann11}).  Here $\sigma^2$ is the true error variance, and $k$ is a constant that does not depend on $n$ or $p$.  These bounds are commonly termed ``oracle inequalities'' since, apart from the $\log(p)$ term and the constant, they equal the loss expected if an oracle told us the true set of predictors and we fit least squares.  In light of this connection, it is commonly noted in the literature that the ``$\log(p)$-factor is the price to pay by not knowing the active set'' \citep{buhlmann13} and ``it is also known that one cannot, in general, hope for a better result'' \citep{candes09}.  Under certain assumptions and an appropriate control of the number of predictor variables, these bounds establish $l_2$-loss consistency in the sense that the $l_2$-loss will tend to zero asymptotically.  Similar upper bounds exist for the expected value of the loss \citep{bunea07a} as well as lower bounds when $\mX$ is non-singular \citep{chatterjee14}. \citet{bunea06} and \citet{bunea07b} further established bounds on the loss for random designs and 
\citet{thrampoulidis2015} studied the asymptotic behavior of the normalized squared error of the Lasso when $p \to \infty$ and $\sigma \to 0$ under the assumption of a Gaussian design matrix.  In related work on the predictive performance, \citet{greenshtein04} and \citet{greenshtein06} also studied the ``persistence'' of the Lasso estimator and showed that the difference between the expected prediction error of the Lasso estimator at a particular deterministic value of $\lambda$ and the optimal estimator converges to zero in probability.  Thus, the ``Lasso achieves a squared error that is not far from what could be achieved if the true sparsity pattern were known'' \citep{vidaurre13}.

Unfortunately, there is a disconnect between these theoretical results and the way that the Lasso is implemented in practice.  In practice $\lambda$ is not taken to be a deterministic value, but rather it is selected using an information criterion such as Akaike's information criterion ($AIC$; \citealp{akaike73}), the corrected $AIC$ ($AIC_c$; \citealp{hurvich89}), the Bayesian information criterion ($BIC$; \citealp{schwarz78}), or Generalized cross-validation ($GCV$; \citealp{craven78}) or by using ($k$-fold) cross-validation ($CV$) (see, e.g., \citealp{fan01}, \citealp{leng06}, \citealp{zou07}, \citealp{yu13}, \citealp{flynn13}, and \citealp{homrighausen14}).  Since the existing theoretical results do not apply to a data-dependent choice of $\lambda$ \citep{chatterjee14}, it is not clear how well the oracle inequalities represent the performance of the Lasso in practice.

This motivates us to study the behavior of the loss at a data-dependent choice of the regularization parameter.  We define the random variable $\lambda_p^* = \argmin_{\lambda} L_p(\lambda)$ to be the optimal (infeasible) choice
of $\lambda$ that minimizes the loss function over the regularization path.  In what follows, we focus on the loss of the Lasso evaluated at $\lambda_p^*$.
This selector provides information about the performance of the method in an absolute sense, and it represents the ultimate goal for any model selection procedure designed for prediction.

By the definition of the optimal loss, the oracle inequalities in the literature also apply to $L_p(\lambda_p^*)$.
It is therefore tempting to use the oracle inequalities in the literature to describe the behavior of the optimal loss.  The work on persistency has also led to conclusions such as ``there is `asymptotically no harm' in introducing many more explanatory variables than observations,'' \citep{greenshtein04} and that ``in some `asymptotic sense', when assuming a sparsity condition, there is no loss in letting [$p$] be much larger than $n$'' \citep{greenshtein06}.  More generally, when working in high-dimensional settings these results are interpreted to imply that ``having too many components does not degrade forecast accuracy'' \citep{hyndman13} and ``it will not hurt to include more variables'' \citep{lin11}.  However, it is important to remember that the existing theoretical results are based on inequalities, not equalities, so they do not necessarily describe the behavior of the optimal loss or the cost of working in high-dimensional settings.  To our knowledge, this is the first explicit study of the sensitivity of the best-case predictive performance to the number of predictor variables.

The remainder of this paper is organized as follows.  Section~\ref{S:theory} presents some theoretical results on the behavior of the Lasso based on a data-dependent choice of $\lambda$ and proves that the best-case predictive performance can deteriorate as the number of predictor variables is increased, in the sense that best-case performance worsens as superfluous variables are added to the set of predictors.  In particular, under the assumption of a sparse true model and orthonormal predictors, we prove that the probability of deterioration is non-zero.  In the special
case where there is only one true predictor, we further prove that the probability of deterioration
can be arbitrarily close to one for a sufficiently high signal to noise ratio and sufficiently large $p$, and that the expected amount of deterioration is infinite.
Section~\ref{S:empirical} investigates the amount of deterioration empirically and shows that it can be much worse than one might expect from looking at the loss bounds in the literature.  Section~\ref{S:data} presents an analysis of HIV data using the Lasso and exemplifies the occurrence of deterioration in practice.  Finally, Section~\ref{S:concl} presents some final remarks and areas for future research.  The appendix includes some additional technical and simulation results.

\section{Theoretical Results}\label{S:theory}
Here we consider a simple framework for which there exists an exact solution for the Lasso estimator.
We assume that
\[
\vy = \mX \vbeta_0 + \varepsilon
\]
where $\vy$ is the $n \times 1$ response vector, $\mX$ is a $n \times p$ matrix of deterministic predictors such that $\mX^T \mX = \mI$ (the $p \times p$ identity matrix), $\vbeta_0 = (\beta_1, \ldots, \beta_p)^T$ is the $p \times 1$ vector of true unknown coefficients, and $\boldsymbol \varepsilon$ is a $n \times 1$ noise vector where $\varepsilon_i \stackrel[]{iid}{\sim} N( 0, \sigma^2)$.  Under the orthonormality assumption, we require $p \leq n$.

We define $p_0$ to be the number of non-zero true coefficients, where $1 \leq p_0 \leq p$.  
Without loss of generality, we assume that $\beta_j \neq 0$ for $1 \leq j \leq p_0$ and $\beta_j = 0$ for $p_0 < j \leq p$.
We further assume that there is no intercept.  

By construction $\vz = \mX^T \vy$ is the vector of the least squares-estimated coefficients based on the full model.  It follows that
the $z_j$'s are independent for all $1 \leq j \leq p$, and that
\begin{equation}\label{e:z1-dist}
z_j \sim N( \beta_j , \sigma^2)
\end{equation}
for $1 \leq j \leq p_0$ and 
\begin{equation}\label{e:zj-dist}
z_j \stackrel[]{iid}{\sim} N( 0 , \sigma^2)
\end{equation}
for $p_0 < j \leq p$.  For a given $\lambda$, the Lasso estimated coefficients are
\[
\hat{\beta}_{\lambda j} = \text{sgn}(z_j) ( |z_j| - \lambda)_+
\]
for $j=1,\ldots,p \;$ \citep{fan01}.  
We use $L_p (\lambda)$ to measure the performance of this estimator. Under our set-up,
\begin{equation}\label{loss}
L_{p} (\lambda) = \frac{1}{n} \sum_{j = 1}^{p_0} (\beta_j - \hat{\beta}_{\lambda j})^2
                    + \frac{1}{n}\sum_{j = p_0 + 1}^{p} \hat{\beta}^2_{\lambda j}.
\end{equation}

We wish to study the sensitivity of the Lasso to the number of predictor variables and to investigate the occurrence of deterioration in practice.
Recall that deterioration is defined to be the worsening of best-case performance as superfluous variables are added to the set of predictors.
Thus, 
deterioration occurs when the optimal loss ratio
\[
\frac{ L_p(\lambda^*_{p}) } { L_{p_0}(\lambda^*_{p_0}) } > 1
\]
for $p > p_0$.  

In what follows, we establish that the best case predictive performance of the Lasso deteriorates as $p$ increases with non-zero probability.
For ease of presentation, the proofs for the technical results in this section are presented in Appendix~\ref{S:tech}.

\begin{thm}\label{T:prob-det-s-sparse}
For all $1 \leq p_0 < p \leq n$,
\begin{equation}\label{e:det-lower-bound}
\Pr \left( \frac{L_{p}(\lambda_p^*)}{L_{p_0}(\lambda_{p_0}^*)} > 1 \right)
> 0.
\end{equation}
\end{thm}

To prove Theorem~\ref{T:prob-det-s-sparse} we make use of the following lemma, which establishes the conditions under which deterioration occurs.

\begin{lemma}\label{l:s-sparse}
For all $1 \leq p_0 < p \leq n$,
\[
\frac{L_p(\lambda^*_p)}{L_{p_0}(\lambda^*_{p_0})} > 1
\]
if and only if
\[
\lambda^*_{p_0} \leq \max_{1 \leq j \leq p_0} |z_j| 
\]
and
\[
\lambda^*_{p_0} \leq \max_{p_0 < j \leq p} |z_j|.
\]
\end{lemma}

To understand the results of Lemma~\ref{l:s-sparse}, first note that for all $p > 0$,
$L_p(\lambda_p^*) \leq \frac{1}{n} \sum_{j = 1}^{p_0} \beta^2_j$, because there always exists a $\lambda$ such that all of the estimated coefficients are shrunk to zero.  Thus, no deterioration occurs in the extreme case where $\lambda_{p_0}^*$ is equal to such a value.  In particular, this occurs if $\lambda_{p_0}^* \geq \max_{1 \leq j \leq p_0} |z_j|$.
Outside of this case,
the optimal loss will deteriorate if we cannot set the estimated coefficients for the extraneous predictors equal to zero without imposing more shrinkage on the
estimated coefficients for the true predictors.  This occurs if $\lambda^*_{p_0} > \max_{p_0 < j \leq p} |z_j|$.

As Lemma~\ref{l:s-sparse} implies, it is possible to establish stronger results about the
probability of deterioration when the behavior of $\lambda^*_{p_0}$
is known. In the remainder
of this section we establish theoretical results in the case where $p_0 = 1$, and in
Appendix~\ref{app:s-2det} we provide results in the case where $p_0 = 2$. In both cases, our results
demonstrate that deterioration occurs with probability arbitrarily close to one for
an appropriately high signal to noise ratio and large p.

In the special case where $p_0 = 1$, it is further possible to derive a simple exact expression for the probability of deterioration.  

\begin{thm}\label{T:prob-det}
For $p_0 = 1$ and for all $1 < p \leq n$,
\begin{equation}\label{e:theorem-result}
\Pr \left( \frac{L_{p}(\lambda_p^*)}{L_{p_0}(\lambda_{p_0}^*)} > 1 \right)
= \Phi \left(\frac{|\beta_1|}{\sigma}\right) - \frac{1}{2p},
\end{equation}
where $\Phi(\cdot)$ is the cumulative distribution function of a standard normal random variable.
\end{thm}

In Appendix~\ref{S:tech}, when $p_0 = 1$, we establish that $nL_p(\lambda^*_p) = \beta_1^2$ for all $p > 0$
if the sign of $z_1$ is incorrect.  This means that no deterioration occurs in this case.  With this result in place, the two terms on the right-hand side of equation~(\ref{e:theorem-result})
can be explained intuitively.  
The first term reflects the increasing
likelihood that the sign of $z_1$ is correct as the signal-to-noise ratio increases, and the second term reflects the
decreasing probability of no deterioration in this case as $p$ increases.  This result establishes that deterioration occurs with probability arbitrarily close to one for an appropriately high signal to noise ratio and large $p$ when $p_0 = 1$, and the following theorem establishes that the expected amount of deterioration is infinite.

\begin{thm}\label{T:exp-det}
For $p_0 = 1$ and for all $1 < p \leq n$,
\[
\E \left( \frac{L_{p}(\lambda_p^*)}{L_{p_0}(\lambda_{p_0}^*)} \right)
= \infty.
\]
\end{thm}

The result of Theorem~\ref{T:exp-det} follows from the fact that the case where $L_{p_0}(\lambda_{p_0}^*) = 0$ and $L_{p}(\lambda_p^*) > 0$ occurs with non-zero probability when $p_0 = 1$.
We further investigate the amount of deterioration in the more general $p_0$-sparse case using simulations in Section~\ref{S:empirical}.

As an alternative to loss, performance could also be measured based on Mean Squared Error (MSE).  
Under the assumption of a deterministic design matrix,
\[
MSE_p(\lambda) = E^*\left(\frac{||\vy^* - \hat{\vmu}_\lambda||^2}{n}\right)
	= \frac{||\vmu - \hat{\vmu}_\lambda||^2}{n} + \frac{\sigma^2}{n}
	= L_p(\lambda) + \frac{\sigma^2}{n},
\]
where $\vy^*$ is from an independent test set and the expectation $E^*$ is taken with respect to this independent test set.
Thus, Theorems~\ref{T:prob-det-s-sparse}-\ref{T:prob-det} also apply to MSE. Since MSE also includes the error variance, 
the relative deterioration of MSE is expected to be less
than that of loss when using the one correct predictor.  We discuss this further in our real data application in Section~\ref{S:data} where we study deterioration in average squared prediction error.

\textbf{Example.} To demonstrate the implications of Theorem~\ref{T:prob-det}, consider an \\* ANOVA model based on an orthonormal regression matrix.  Specifically, assume that we have $p$ binary predictor variables, each of which is coded using effects coding, and a balanced design with an equal number of observations falling into each of the $2^p$ combinations.  If we scale these predictors to have unit variance, then an ANOVA model on only the main effects is equivalent to a regression on these predictors.  Similarly, if we consider all pairwise products and then standardize, a regression including them as well as the main effects is equivalent to an ANOVA with all two-way interactions.  We can continue to add higher-order interactions in a similar manner, where a model with all $k$-way interactions includes $\sum_{i=1}^k \binom{p}{i}$ predictors.

Assume that only the main effect of the first predictor has a nonzero effect, $\beta_1=3$, and that $\sigma = 1$.  Then applying the result of Theorem~\ref{T:prob-det}, Table~\ref{tab:prob-det} shows that the probability of deterioration can be close to one for even a moderate number of predictor variables.

\begin{table}
\caption{\label{tab:prob-det}The probability of deterioration when only the main effect of the first predictor has a nonzero effect, $\beta_1=3$, $\sigma=1$, and higher order interactions are included.}
\centering
\fbox{%
\begin{tabular}{l ccccc}
& \multicolumn{5}{c}{Probability of Deterioration}\\
Model                   & $p=2$       & $p=4$       & $p=6$       & $p=8$       & $p=10$\\
Main Effects            &0.7487 &    0.8737      & 0.9154    & 0.9362    & 0.9487 \\
Two-Way Interactions    &0.8362 &    0.9487      & 0.9749    & 0.9848    & 0.9896 \\
Three-Way Interactions  &$-$      &    0.9602      & 0.9865    & 0.9933    & 0.9958 \\
Four-Way Interactions   &$-$      &    0.9630      & 0.9898    & 0.9956    & 0.9974 \\
\end{tabular}}
\end{table}

\section{Empirical Study}\label{S:empirical}

This section empirically investigates the cost of not knowing the true set of predictors when working with high-dimensional data.  We assume that $\vy$ is generated by the model in (\ref{e:dgp}).  The Lasso regressions are fit using the R {\tt  glmnet} package \citep{friedman10}.  We use the default package settings and include an intercept in the model.  We consider two simulation set-ups.  The first studies the performance of the Lasso when the columns of $\mX$ are trigonometric predictors.  Since these predictors are orthogonal, this setting requires $p<n$.  To allow for situations with $p>n$, we also study the case where the columns of $\mX$ are independent standard normals.

The main goal of our simulations is to understand the behavior of the infeasible optimal loss for the Lasso
as $p$ and $n$ vary.  To measure the deterioration in optimal loss we consider the optimal loss ratio
\begin{equation}\label{e:opt-loss-ratio}
\frac{L_{p}(\lambda_p^*)}{L_{p_0}(\lambda_{p_0}^*)},
\end{equation}
which compares the minimum loss based on $p$ predictors to the minimum loss based on the true set of $p_0$ predictors.  These $p_0$ predictors have nonzero coefficients.  All other coefficients are zero.  Here $p_0 < p$ and the $p_0$ true predictors are always a subset of the $p$ predictors.  We focus on cases where $p$ is large or grows with $n$ in order to be consistent with high-dimensional frameworks.

By the definition of $\lambda_p^*$, the oracle inequalities in the literature also apply to $L_p(\lambda_p^*)$.
In what follows, we compare the empirical performance of the optimal loss (computed over the default grid of $\lambda$ values) to two established bounds.
First, by applying Corollary 6.2 in \citet{buhlmann11},
\begin{equation}\label{e:bound1}
L_p(\lambda_p^*) \leq 64 \sigma^2 p_0 \frac{t^2 + 2\log(p)}{n \, \psi^2_0}
\end{equation}
with probability greater than $1-2e^{-t^2/2}$ for any constant $t > 0$, where $\psi_0$ is a constant that satisfies a compatibility condition.  This condition places a restriction on the minimum eigenvalue of $\mX^T \mX/n$ for a restricted set of coefficients and it's sufficient to take $\psi_0=1$ for an orthogonal design matrix.
Second, by Theorem 6.2 in \citet{bickel10},
\begin{equation}\label{e:bound2}
L_p(\lambda_p^*) \leq 16 A^2 \sigma^2 p_0 \frac{\log(p)}{n \, \kappa^2}
\end{equation}
with probability at least $1-p^{1-A^2/8}$ for any constant $A > 0$, where $\kappa$ is a constant tied to a restricted eigenvalue assumption.  For orthogonal predictors, $\kappa=1$.  In the simulations, $t$ and $A$ are both set so that the bounds hold with at least 95 percent probability.  Since these bounds also depend on $p$, we study if the deterioration in optimal loss is adequately predicted by these bounds by comparing the observed optimal loss ratio to the 
loss bound ratio.  Here we define the loss bound ratio to be the ratio that compares each bound based on $p$ predictors to the corresponding bound based on $p_0$ predictors.  The results based on (\ref{e:bound1}) and (\ref{e:bound2}) are similar in the simulation examples in Sections~\ref{S:orthog-pred} and~\ref{S:indep-pred}, so only the results for (\ref{e:bound1}) are reported.

In addition to the infeasible optimal loss, we also consider the performance of the Lasso when tuned using 10-fold CV.  For each simulation, we denote the CV-selected $\lambda$ by $\lambda_p^{CV}$ with corresponding loss $L_p(\lambda_p^{CV})$.  The CV loss ratio is then
computed as
\[
\frac{L_p(\lambda_p^{CV})}{L_{p_0}(\lambda_{p_0}^{CV})}.
\]
Although the bounds in equations~(\ref{e:bound1}) and~(\ref{e:bound2}) are not guaranteed to hold for $\lambda^{CV}_p$,
we compare the observed CV loss ratios to the loss bound ratios to determine how well they predict the Lasso's performance in practice.

\subsection{Orthogonal Predictors}\label{S:orthog-pred}
Define the true model to be
\begin{equation}\label{e:sim_dgp}
y_i = 6x_{i,1}+5x_{i,2}+4x_{i,3}+3x_{i,4}+2x_{i,5}+x_{i,6}+\varepsilon_i
\end{equation}
for $i=1,\ldots,n$, where $\varepsilon_i \stackrel[]{iid}{\sim} N(0, \sigma^2)$.  We compare $\sigma^2=4$ and $\sigma^2=400$ in order to
study the impact of varying the signal-to-noise ratio (SNR).  We refer to these cases as ``High SNR'' and ``Low SNR'', respectively.

The columns of $\mX$ are trigonometric predictors defined by
\begin{equation*}
x_{i,2j-1} = \sin \left(\frac{2\pi j}{n} (i-1) \right)
\end{equation*}
and
\begin{equation*}
x_{i,2j} = \cos \left(\frac{2\pi j}{n} (i-1) \right)
\end{equation*}
for $j = 1, \ldots, p/2$ and $i=1, \ldots, n$.  The columns of $\mX$ are orthogonal under this design and the
true model is always included among the candidate models.

\begin{figure}[H]
\centering
\caption{Optimal loss percentiles over 1000 realizations as a function of $p$ for $n=100$ and $p_0=6$.  The number of predictor variables $p$ is varied from $6$ to $100$.  The ``High SNR'' and ``Low SNR'' settings correspond to $\sigma^2=4$ and $\sigma^2=400$, respectively.}
\label{f:loss_nfixed}
\subfigure[High SNR]{
\makebox{\includegraphics[width=6.5cm]{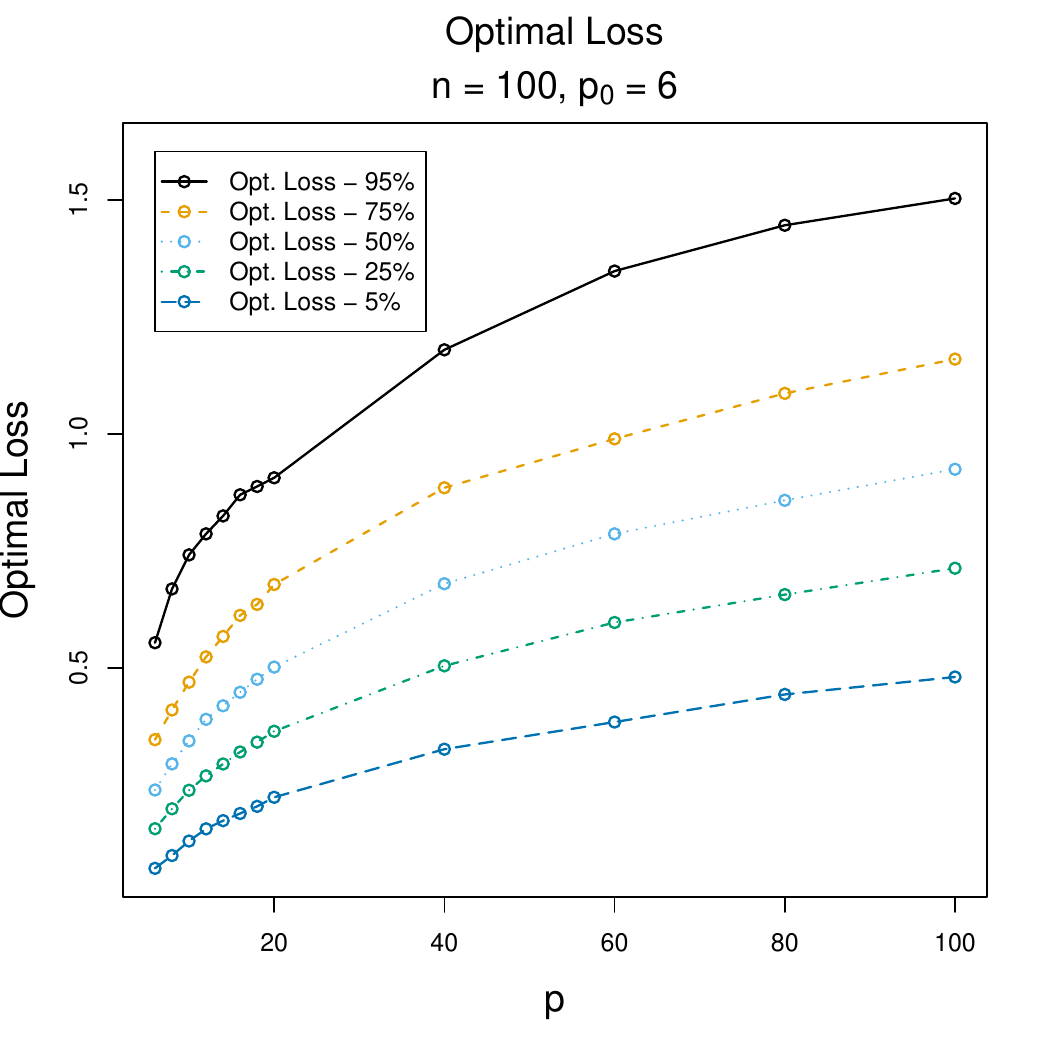}}}
\subfigure[Low SNR]{
\makebox{\includegraphics[width=6.5cm]{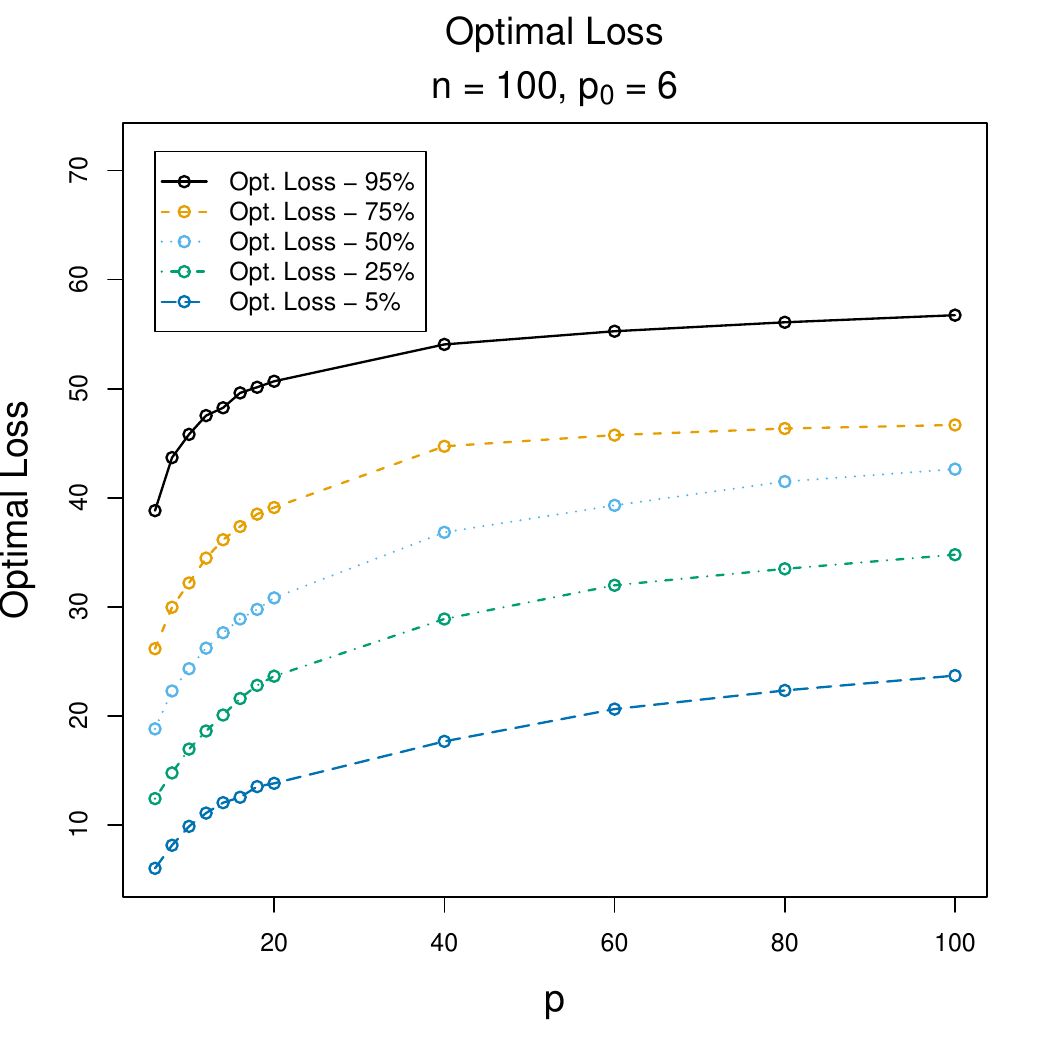}}}
\end{figure}

\begin{figure}[H]
\centering
\caption{Percentiles of the optimal loss ratios over 1000 realizations as a function of $\log(p)$ for $n=100$ and $p_0=6$.  The number of predictor variables $p$ is varied from $6$ to $100$.  The ``High SNR'' and ``Low SNR'' settings correspond to $\sigma^2=4$ and $\sigma^2=400$, respectively.}
\label{f:ratios_nfixed}
\subfigure[High SNR]{
\makebox{\includegraphics[width=6.5cm]{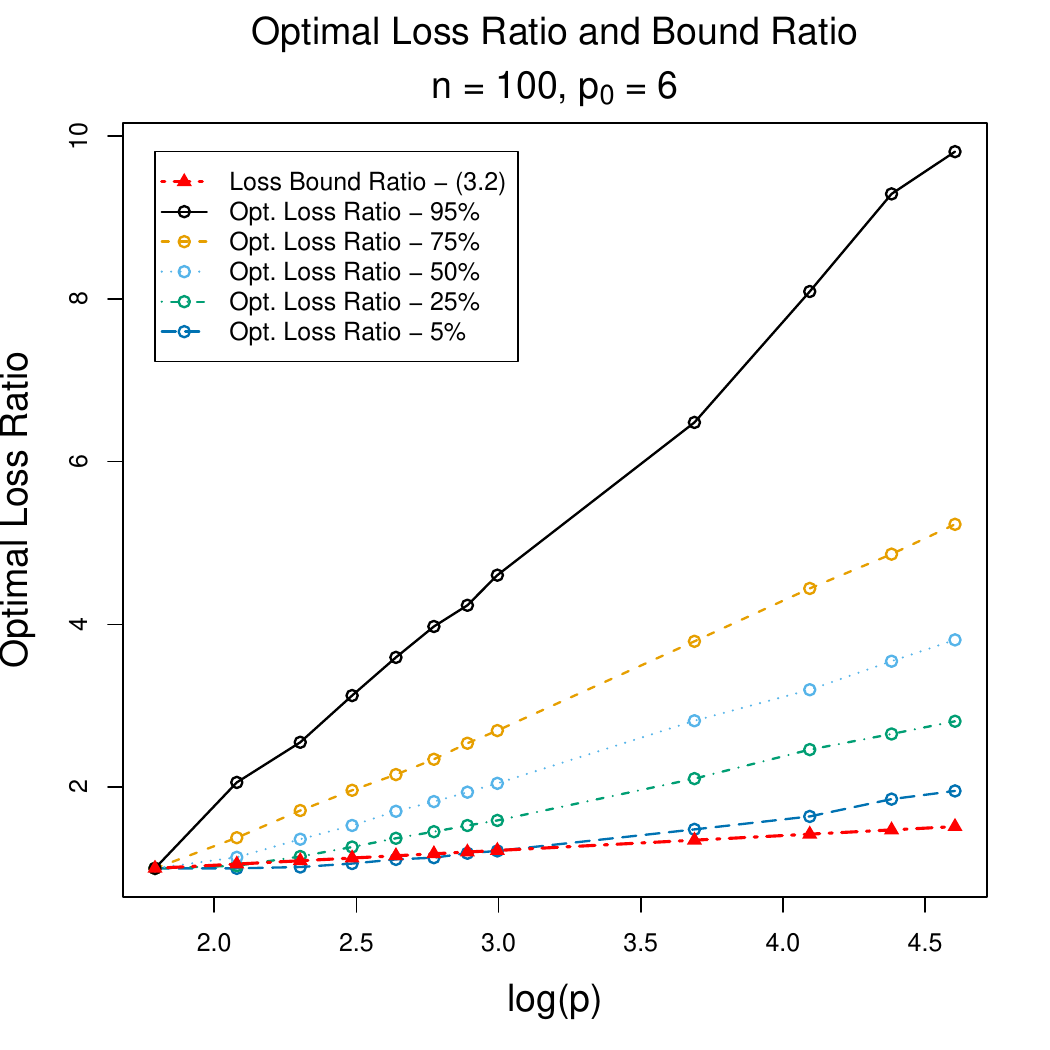}}}
\subfigure[Low SNR]{
\makebox{\includegraphics[width=6.5cm]{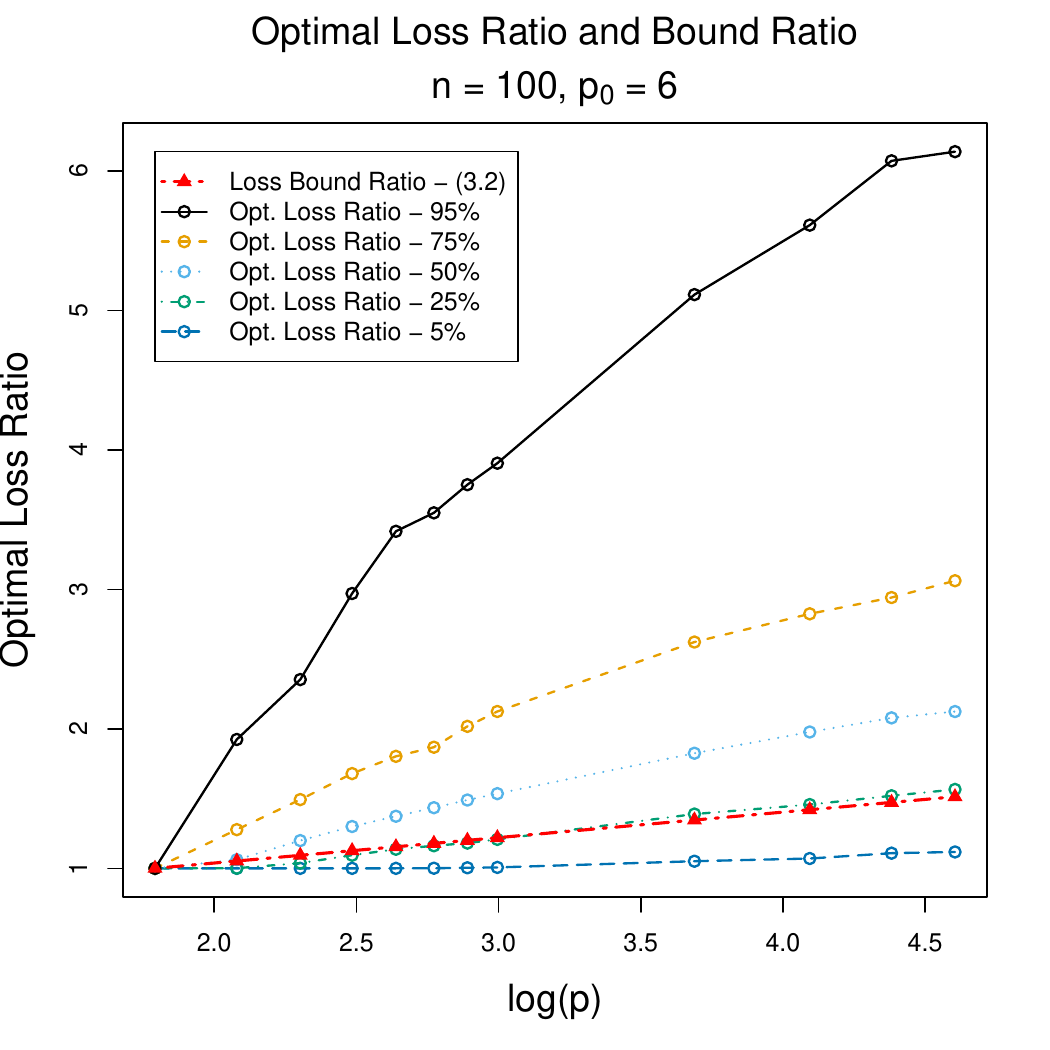}}}
\end{figure}

We first compute the optimal loss, $L_p(\lambda_p^*)$, for varying values of $p$ over 1000 realizations.  Figure~\ref{f:loss_nfixed} plots the percentiles of the optimal losses as a function of $p$.  In both the high and low SNR settings there are signs of deterioration in optimal performance as the number of predictor variables increases, as evidenced by the positive slopes of the percentiles as $p$ increases.  To compare this deterioration to the bounds, Figure~\ref{f:ratios_nfixed} plots the percentiles of the optimal loss ratios over 1000 realizations and the ratio suggested by the loss bound for varying values of $p$.  In both plots, the loss ratios implied by assuming that the bound equals the optimal loss typically under-estimate the observed optimal loss ratio.  Comparing the two plots, the deterioration is worse in the high SNR case.  This is consistent with our theoretical results, which established that we are more likely to observe deterioration when the SNR is high.  When the SNR is low, it is more likely that the optimal loss will equal the loss for $\lambda=\lambda_p^{\max}$, where $\lambda_p^{\max}$ is equal to the value of $\lambda$ that sets all of the $p$ estimated coefficients equal to zero.  When this is the case, no further deterioration can occur when adding more superfluous variables.

Clearly the amount of deterioration is typically far worse than is suggested by the bounds for both choices of the SNR.  For example, looking at the median optimal loss ratio, if we include $n = 100$ predictors in the high SNR case, then the loss bounds suggest we should be about 50 percent worse off than if we knew the true set of predictors, but in actuality we are typically more than 300 percent worse off.  This discrepancy is a consequence of the fact that the bounds are inequalities rather than equalities.

\begin{figure}[H]
\centering
\caption{Ratio of the loss bounds to the observed optimal loss percentiles over 1000 realizations as a function of $p$ for $n=100$.  The ``High SNR'' and ``Low SNR'' settings correspond to $\sigma^2=4$ and $\sigma^2=400$, respectively.}
\label{f:cons_nfixed}
\subfigure[High SNR]{
\resizebox{6.5cm}{!}{\includegraphics{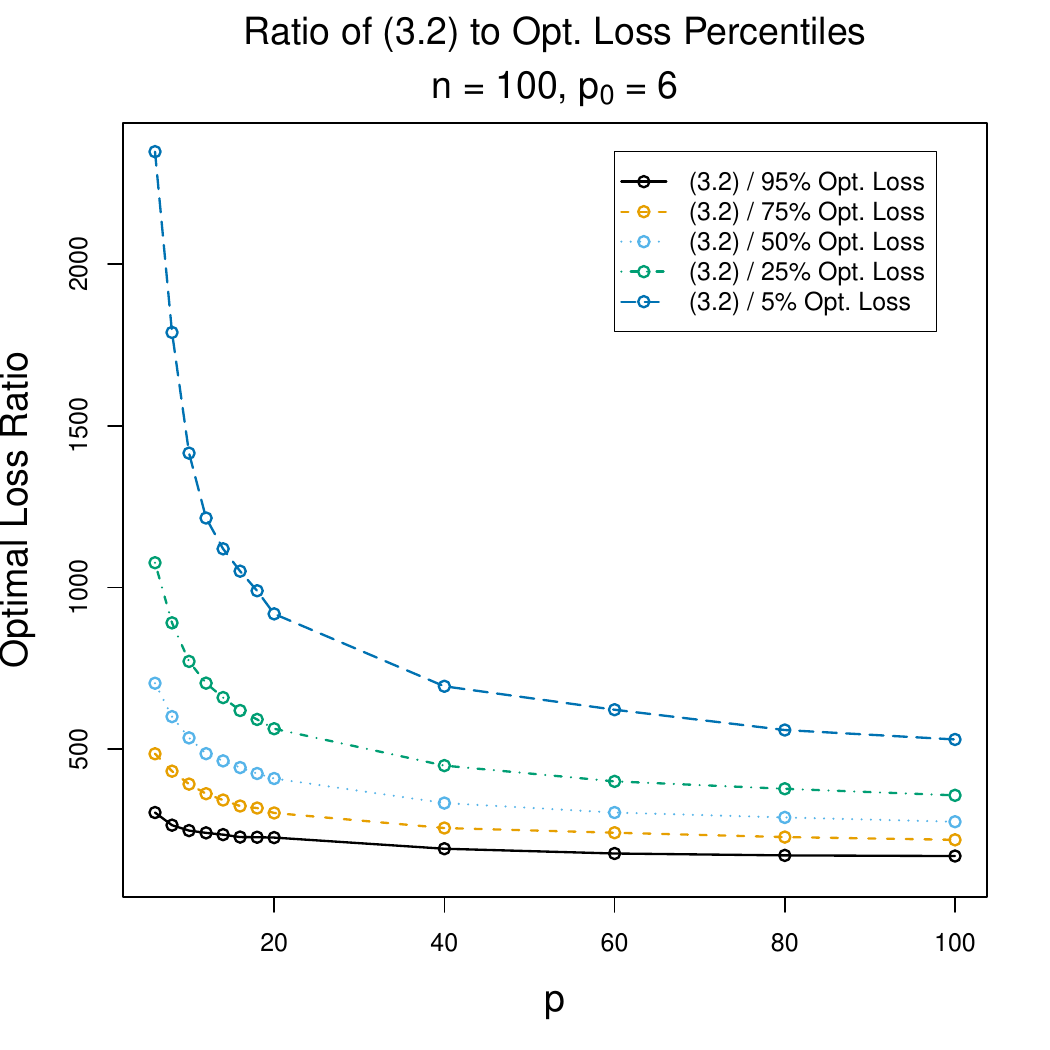}}}
\subfigure[Low SNR]{
\resizebox{6.5cm}{!}{\includegraphics{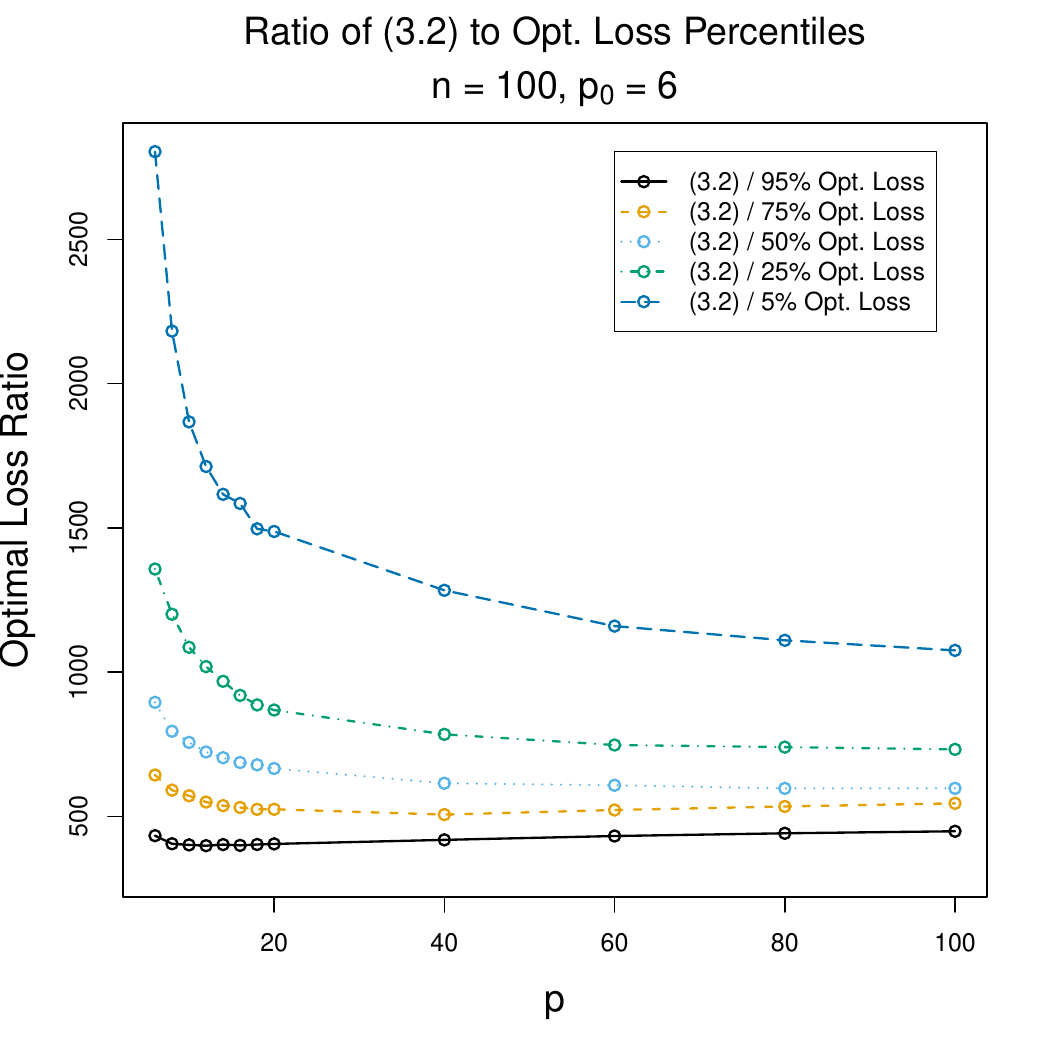}}}
\end{figure}

\begin{figure}[H]
\centering
\caption{Average ratio of $\lambda^0$ to the observed selected $\lambda_p^*$ over 1000 realizations as a function of $p$ for $n=100$ plotted on a log-scale.  The ``High SNR'' and ``Low SNR'' settings correspond to $\sigma^2=4$ and $\sigma^2=400$, respectively.}
\label{f:lambda_nfixed}
\subfigure[High SNR]{
\resizebox{6.5cm}{!}{\includegraphics{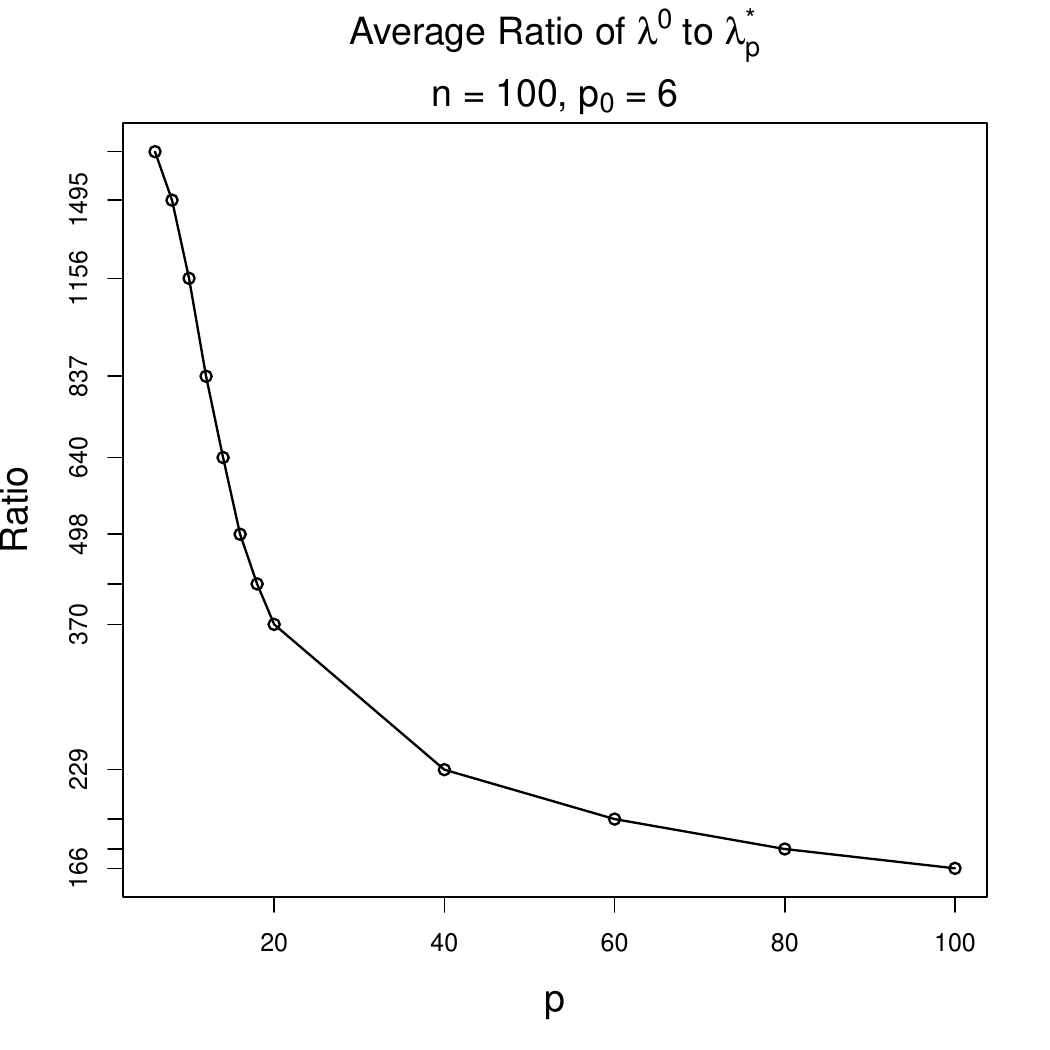}}}
\subfigure[Low SNR]{
\resizebox{6.5cm}{!}{\includegraphics{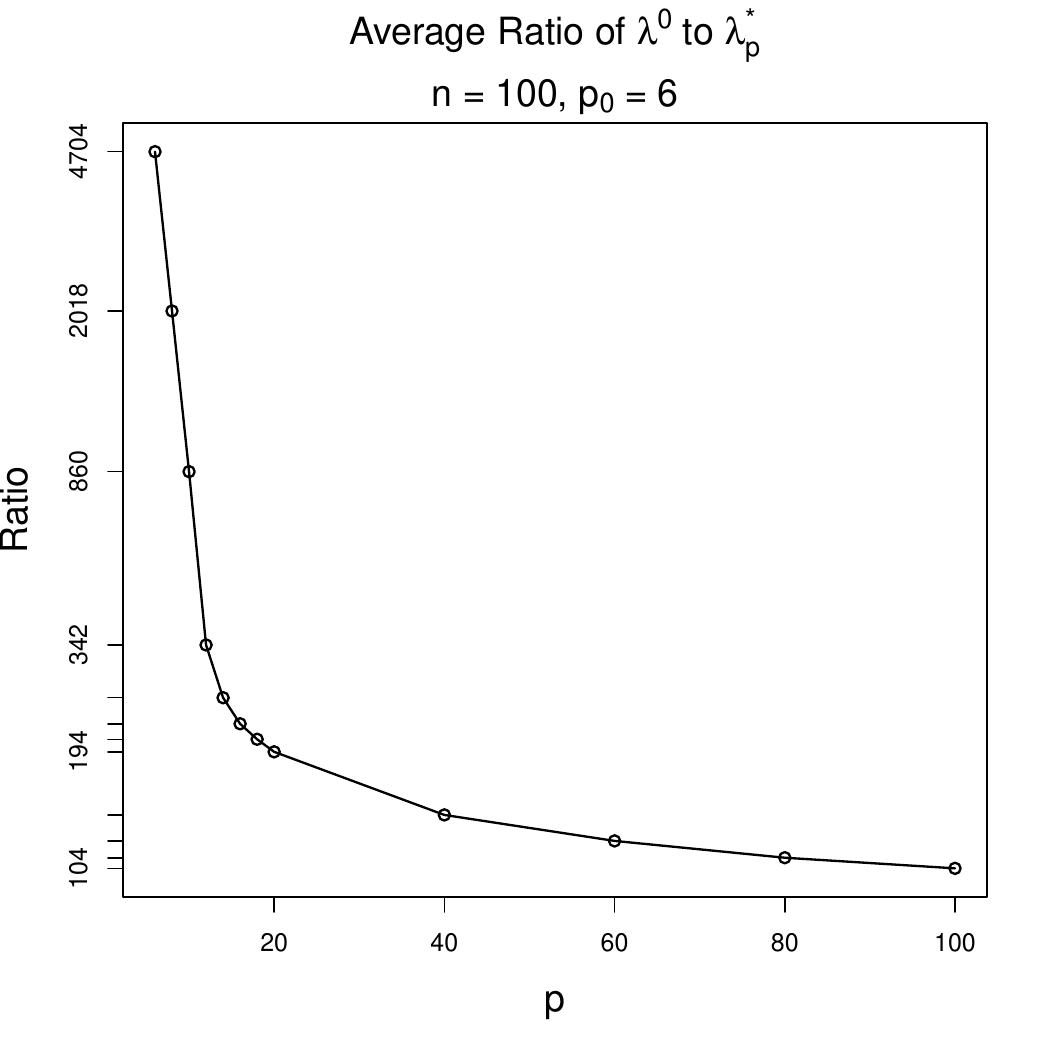}}}
\end{figure}

To emphasize the danger of over-interpreting the bounds, Figure~\ref{f:cons_nfixed} plots the ratio of the bounds to the optimal loss percentiles for varying values of $p$.  These plots suggest that the bounds are overly conservative when compared to the optimal loss and
the degree of conservatism depends on both $p$ and the SNR.
Thus, although the bounds apply, the slope of the optimal loss as a function of $p$ is different than the slope suggested by the bound.  As a result of this behavior, the amount of deterioration in optimal loss can be much worse than the bounds suggest.
To provide further insight, Figure~\ref{f:lambda_nfixed} plots the average ratio of $\lambda^0$ to $\lambda_p^*$ plotted on a log-scale (recall that $\lambda_0$ is the deterministic choice of $\lambda$ used in the oracle inequality~(\ref{e:loss_bound_rough})).  These plots indicate that $\lambda^*_p$ is typically much smaller than $\lambda^0$. 

\begin{figure}[H]
\centering
\caption{Percentiles of the CV loss ratios over 1000 realizations as a function of $\log(p)$ for $n=100$ and $p_0=6$.  The number of predictor variables $p$ is varied from $6$ to $100$.  The ``High SNR'' and ``Low SNR'' settings correspond to $\sigma^2=4$ and $\sigma^2=400$, respectively.}
\label{f:ratios_nfixed_cv}
\subfigure[High SNR]{
\makebox{\includegraphics[width=6.5cm]{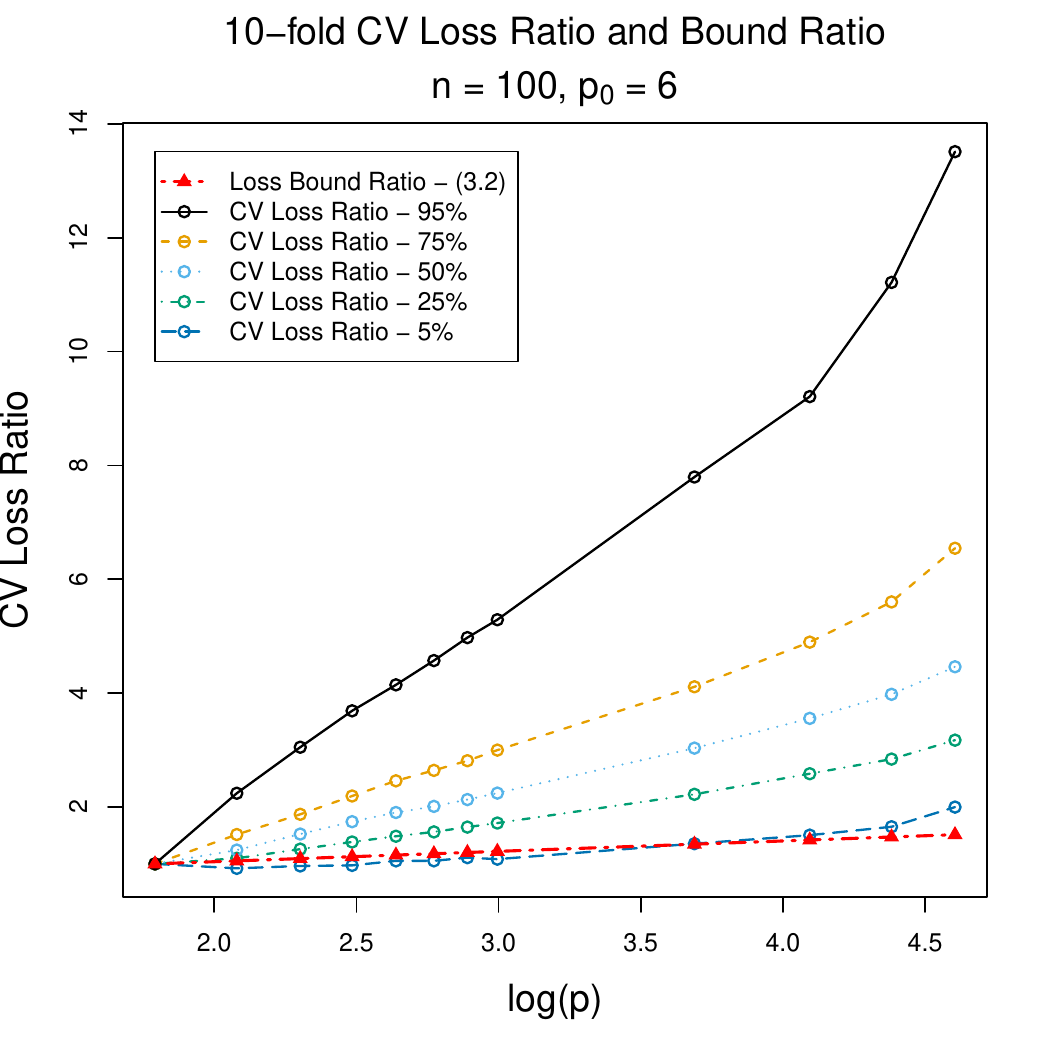}}}
\subfigure[Low SNR]{
\makebox{\includegraphics[width=6.5cm]{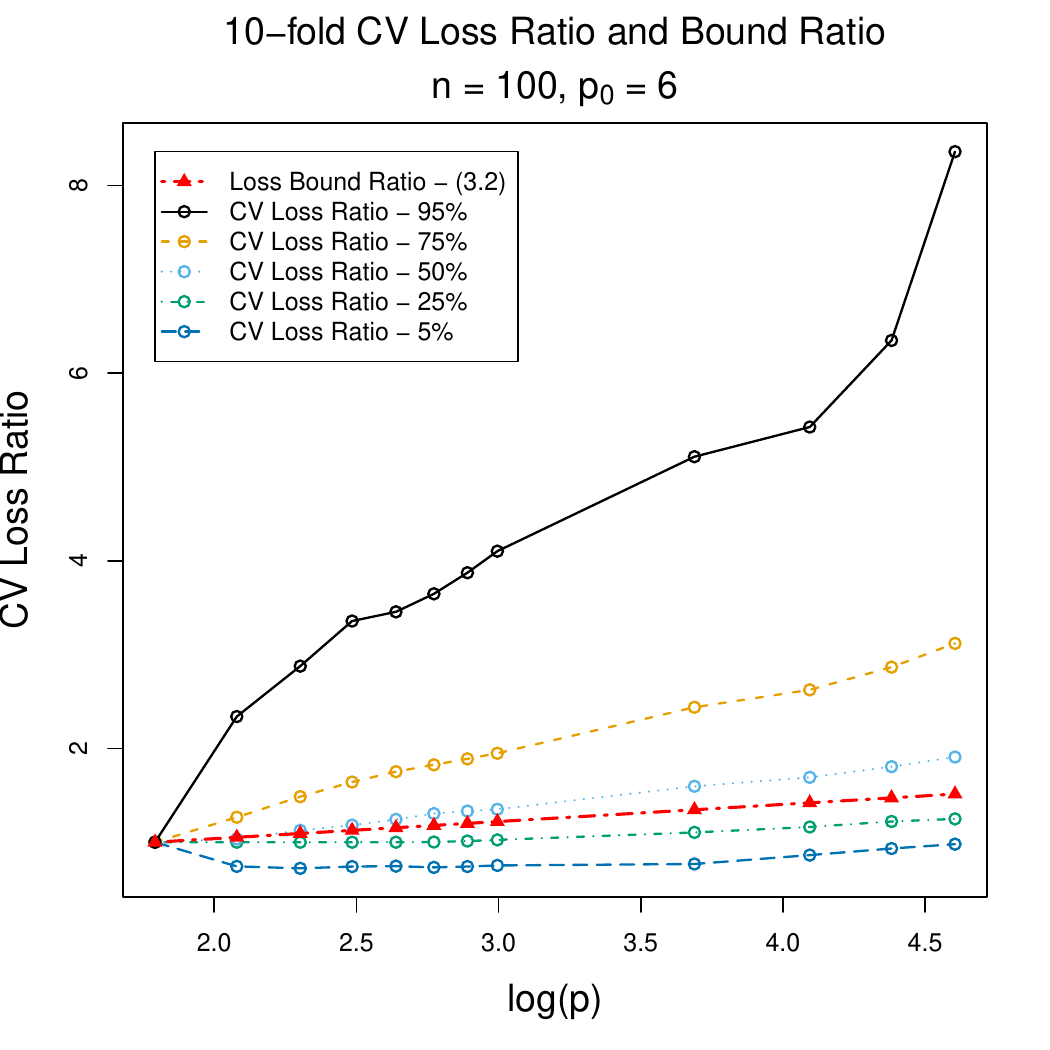}}}
\end{figure}

The optimal selector provides the best-case performance of the Lasso, but it is infeasible in practice.  This motivates us to also study the performance of the Lasso when $\lambda$ is selected in a feasible manner using 10-fold CV.  Figure~\ref{f:ratios_nfixed_cv} compares the CV loss ratios to the bound ratios for varying values of $p$ in the high and low SNR settings.  Similar to the optimal loss, we observe deterioration in the CV loss as $p$ increases that is typically worse than the deterioration suggested by the bounds in both SNR settings.  

\begin{figure}[H]
\centering
\caption{Percentiles of the optimal loss ratios for $p_2=n$ predictors compared to $p_1=2\log(n)$ predictors over 1000 realizations as a function of $n$.  The ``High SNR'' and ``Low SNR'' settings correspond to $\sigma^2=4$ and $\sigma^2=400$, respectively.}
\label{f:ratios_nvaries}
\subfigure[High SNR]{
\resizebox{6.5cm}{!}{\includegraphics{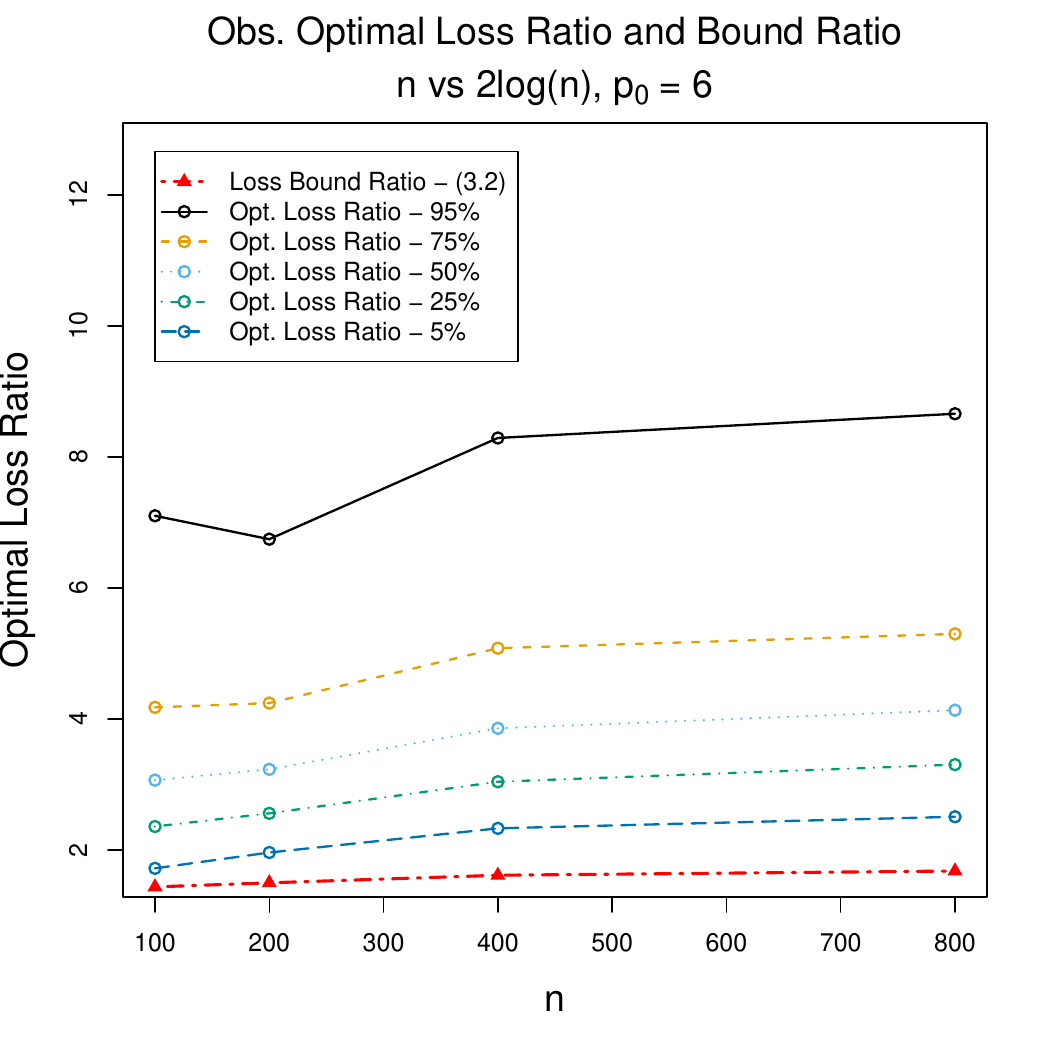}}}
\subfigure[Low SNR]{
\resizebox{6.5cm}{!}{\includegraphics{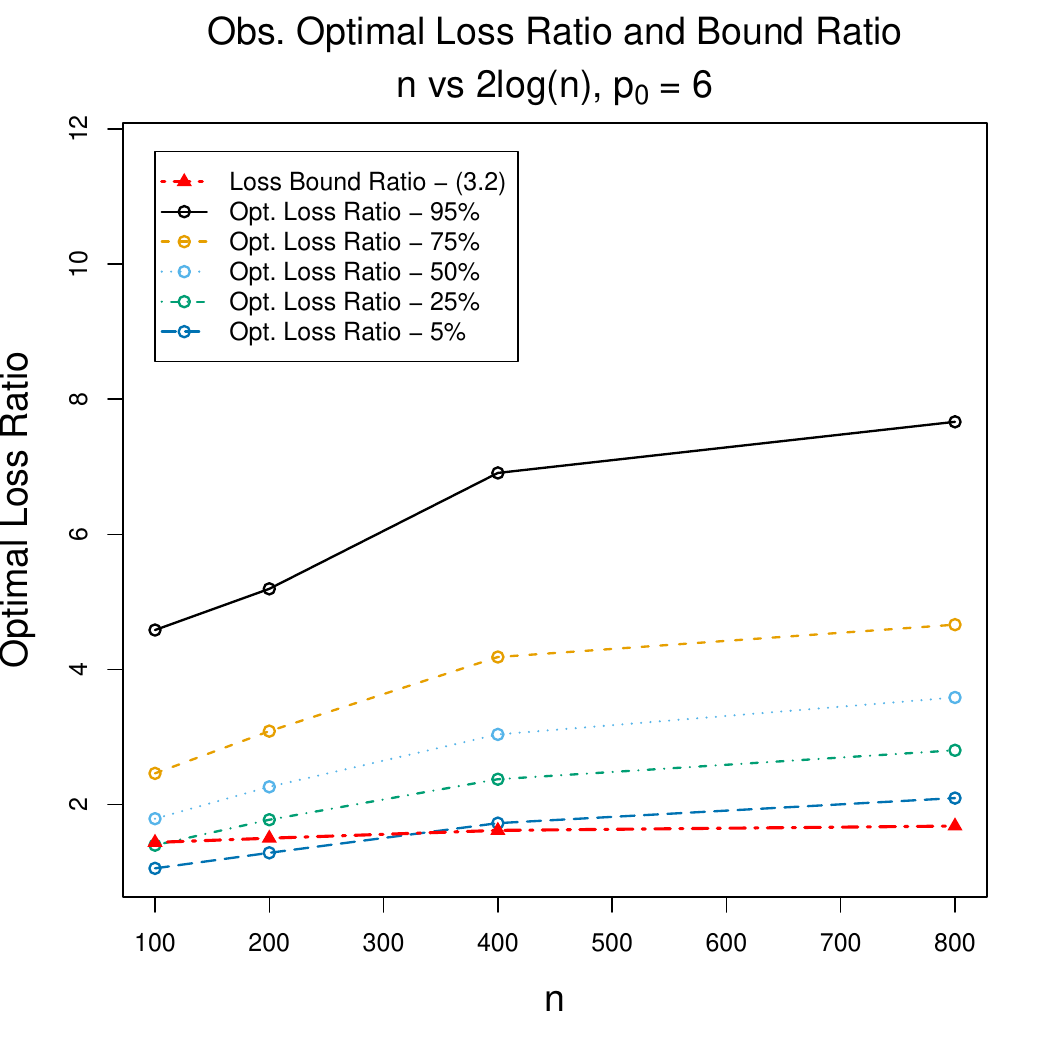}}}
\end{figure}

The results presented thus far suggest that the performance of the Lasso deteriorates for fixed $n$ as $p$ varies.  In order to investigate its behavior when $n$ varies, we compare $p_1 = 2\log(n)$ against $p_2 = n$ and define the optimal loss ratio to be
\[
\frac{L_{p_2}(\lambda_{p_2}^*)}{L_{p_1}(\lambda_{p_1}^*)}.
\]
Under this set-up, $p$ increases as $n$ increases, which is consistent with the standard settings in high-dimensional data analysis.  Figure~\ref{f:ratios_nvaries} compares the percentiles of the optimal loss ratios over 1000 realizations to the optimal loss ratio suggested by the bounds.  These plots suggest that the deterioration persists as $n$ increases, and that the bounds under-predict the observed deterioration.  Since the slopes with respect to $n$ are higher than the bounds imply, this further suggests that the deterioration gets worse for larger samples.

\subsection{Independent Predictors}\label{S:indep-pred}
Here we again assume that $\vy$ is generated from the model given by (\ref{e:sim_dgp}) except in this section the columns of $\mX$ are independent standard normal random variables.  This allows us to consider situations where $p>n$.  This matrix is simulated once and used for all realizations.  We consider both a high and low SNR setting by taking $\sigma^2 = 9$ and $\sigma^2=625$, respectively.

\begin{figure}[H]
\centering
\caption{Percentiles of the optimal and CV loss ratios over 1000 realizations as a function of $\log(p)$ for $n=100$ and $p_0=6$.  The number of predictor variables $p$ is varied from $6$ to $1000$, and the vertical line indicates the point where $p = n$.  The ``High SNR'' and ``Low SNR'' settings correspond to $\sigma^2=9$ and $\sigma^2=625$, respectively.}
\label{f:ratios_pgn}
\subfigure[High SNR, Optimal Selector]{
\resizebox{6.5cm}{!}{\includegraphics{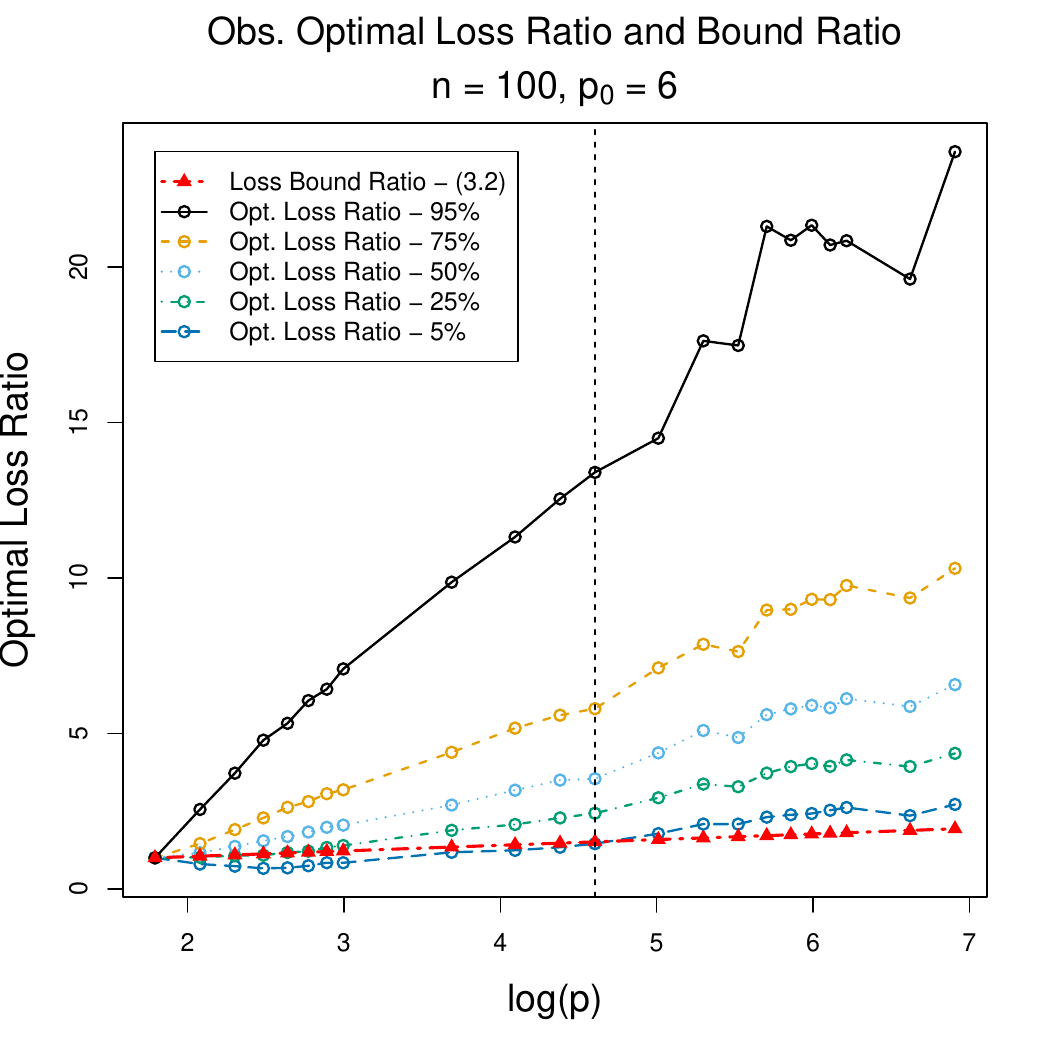}}}
\subfigure[Low SNR, Optimal Selector]{
\resizebox{6.5cm}{!}{\includegraphics{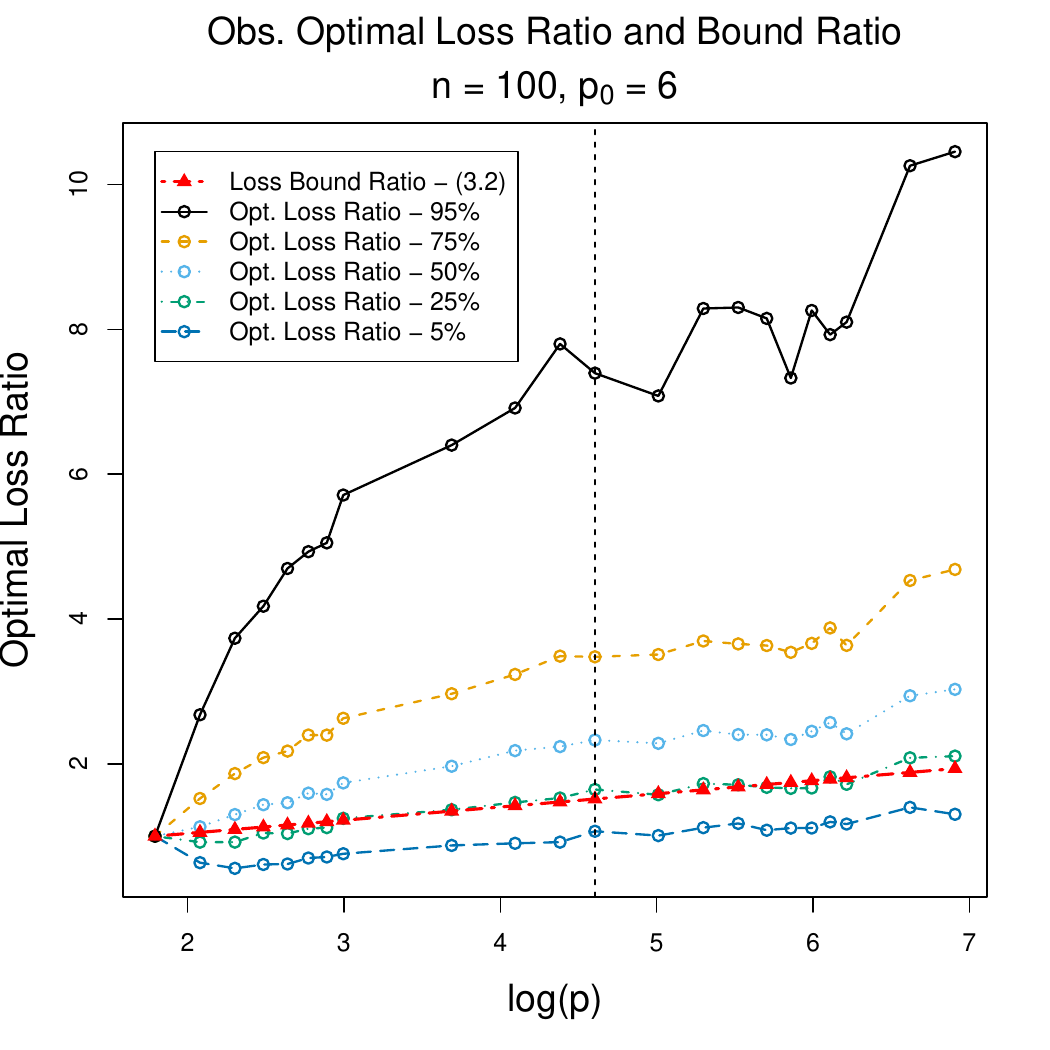}}}
\subfigure[High SNR, 10-fold CV]{
\resizebox{6.5cm}{!}{\includegraphics{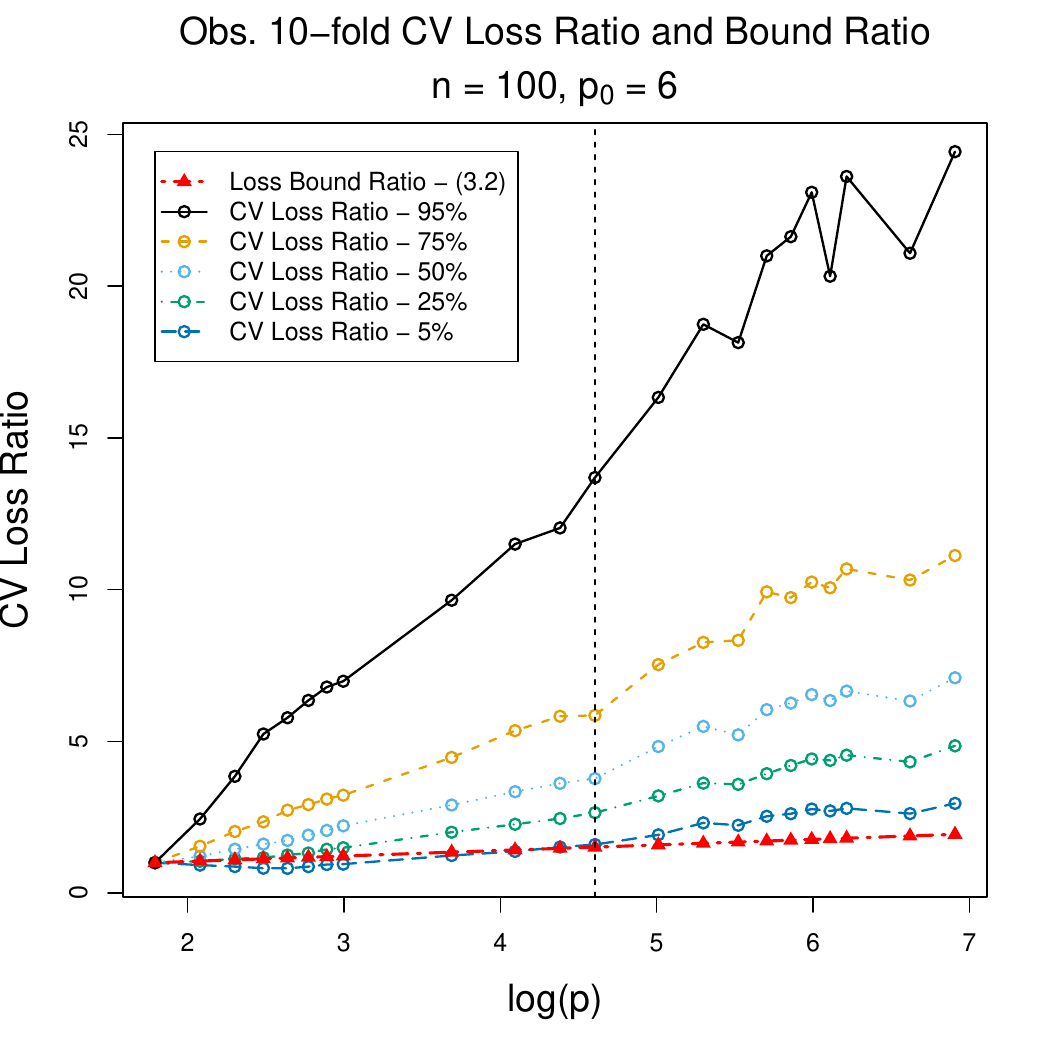}}}
\subfigure[Low SNR, 10-fold CV]{
\resizebox{6.5cm}{!}{\includegraphics{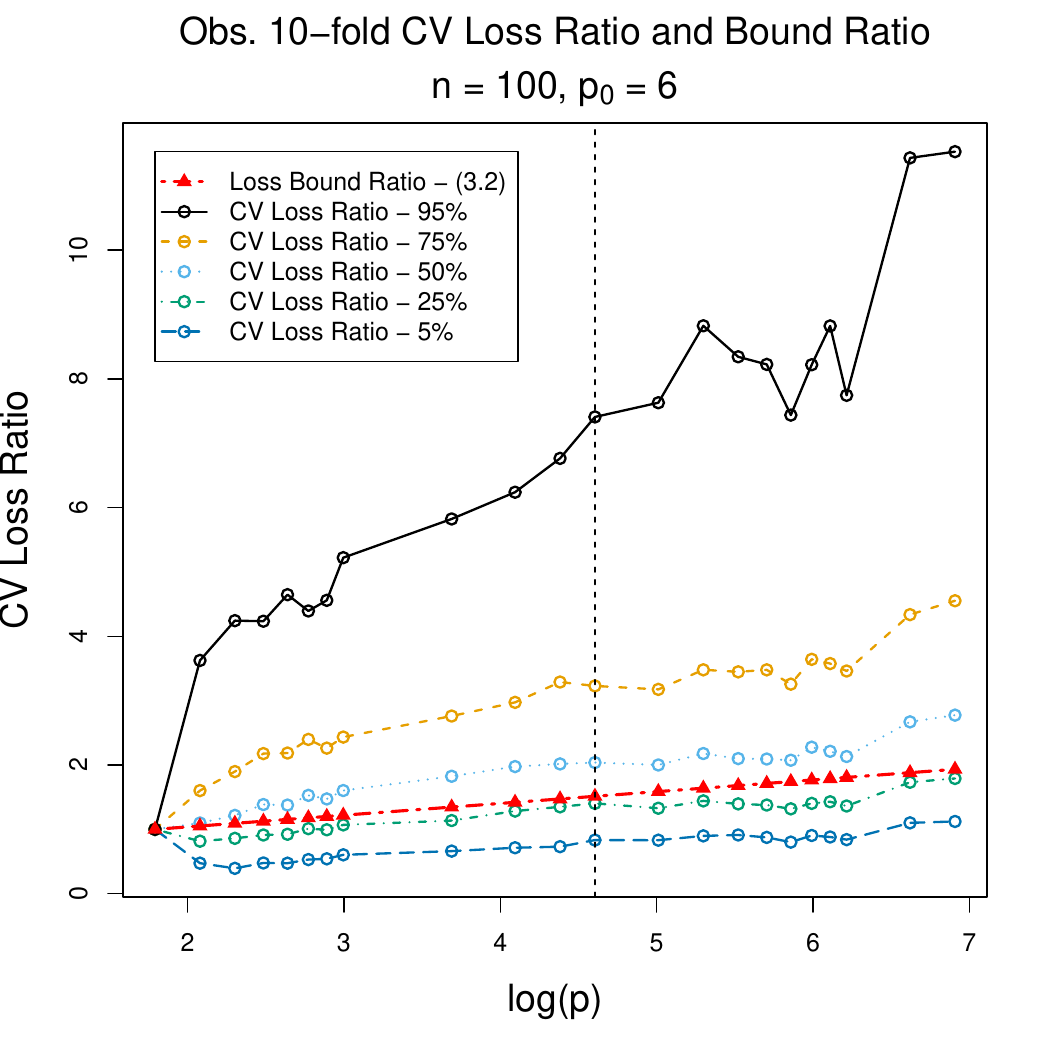}}}
\end{figure}

Figure~\ref{f:ratios_pgn} compares the percentiles of the optimal and CV loss ratios over 1000 realizations to the optimal loss ratio suggested by the bound (\ref{e:bound1}).  We vary $p$ from six to 1000, and denote the point where $p = n$ by the vertical line.  In all four plots, the loss ratios predicted by the bound typically under-estimate the observed optimal and CV loss ratios.  As in the orthogonal design case, these plots show that the bound does not adequately measure the deterioration in performance, and that the optimal and practical performance of the Lasso are sensitive to the number of predictor variables.  These plots further indicate that deterioration occurs when $p>n$, though the deterioration pattern is less well-behaved.

\section{Real Data Analysis}\label{S:data}
In numerous applications it is desirable to model higher-order interactions; however, the inclusion of such interactions can greatly increase the computational burden of a regression analysis.  The Lasso provides a computationally feasible solution to this problem.

As an example of this, \citet{bien13} used the Lasso to investigate the inclusion of all pairwise interactions in the analysis of six HIV-1 drug datasets.  The goal of this analysis was to understand the impact of mutation sites on antiretroviral drug resistance.
These datasets were originally studied by \citet{rhee06} and include a measure of (log) susceptibility for different combinations of mutation sites for each of six nucleoside reverse transcriptase inhibitors.  The number of samples ($n$) and the number of mutation sites ($p$) for each dataset are listed in Table~\ref{tab:HIV-det}.

In their analysis, \citet{bien13} compared the performance of the Lasso with only main effects included in the set of predictors (MEL) to its performance with main effects and all pairwise interactions included (APL).  Although not the focus of their analysis, we show here that this application demonstrates the sensitivity of the procedure to the number of predictor variables, which can result in deteriorating performance in the absence of strong interaction effects.

\begin{table}
\caption{\label{tab:HIV-det}The number of samples and mutation sites in each of the six HIV-1 drug datasets.}
\centering
\fbox{%
\begin{tabular}{lcccccc}
 & \multicolumn{6}{c}{Drug} \\
 & 3TC & ABC & AZT & D4T & DDI & TDF \\
$n$ & 1057 & 1005 & 1067 & 1073 & 1073 & 784 \\
$p$ & 217 & 211 & 218 & 218 & 218 & 216 \\
\end{tabular}}
\end{table}

Since the true data-generating mechanism is unknown, we cannot compute the optimal loss ratios for this example.  As an alternative, to measure deterioration we randomly split the data into a training- and test-set.  We then fit the Lasso using the training-set and evaluate the predictive performance on the test-set by computing the average predictive square error (APSE), which is defined as the average squared error between the values of the dependent variable on the test set and the values predicted by the model fit to the training set.
We then study the APSE ratio, which compares the optimal APSE for APL to the optimal APSE for MEL.
It is important to note that both the numerator and denominator in the APSE ratio include additional terms that depend on the noise term, which are not included in the loss.  Thus, the loss in estimation precision can be less apparent.  To exemplify this, Appendix~\ref{app:mse} studies the optimal APSE ratio in the context of the independent predictors example given in Section~\ref{S:indep-pred}.

\begin{figure}[H]
\centering
\makebox{\includegraphics[width=10cm, trim={2cm 2cm 2cm 2cm}]{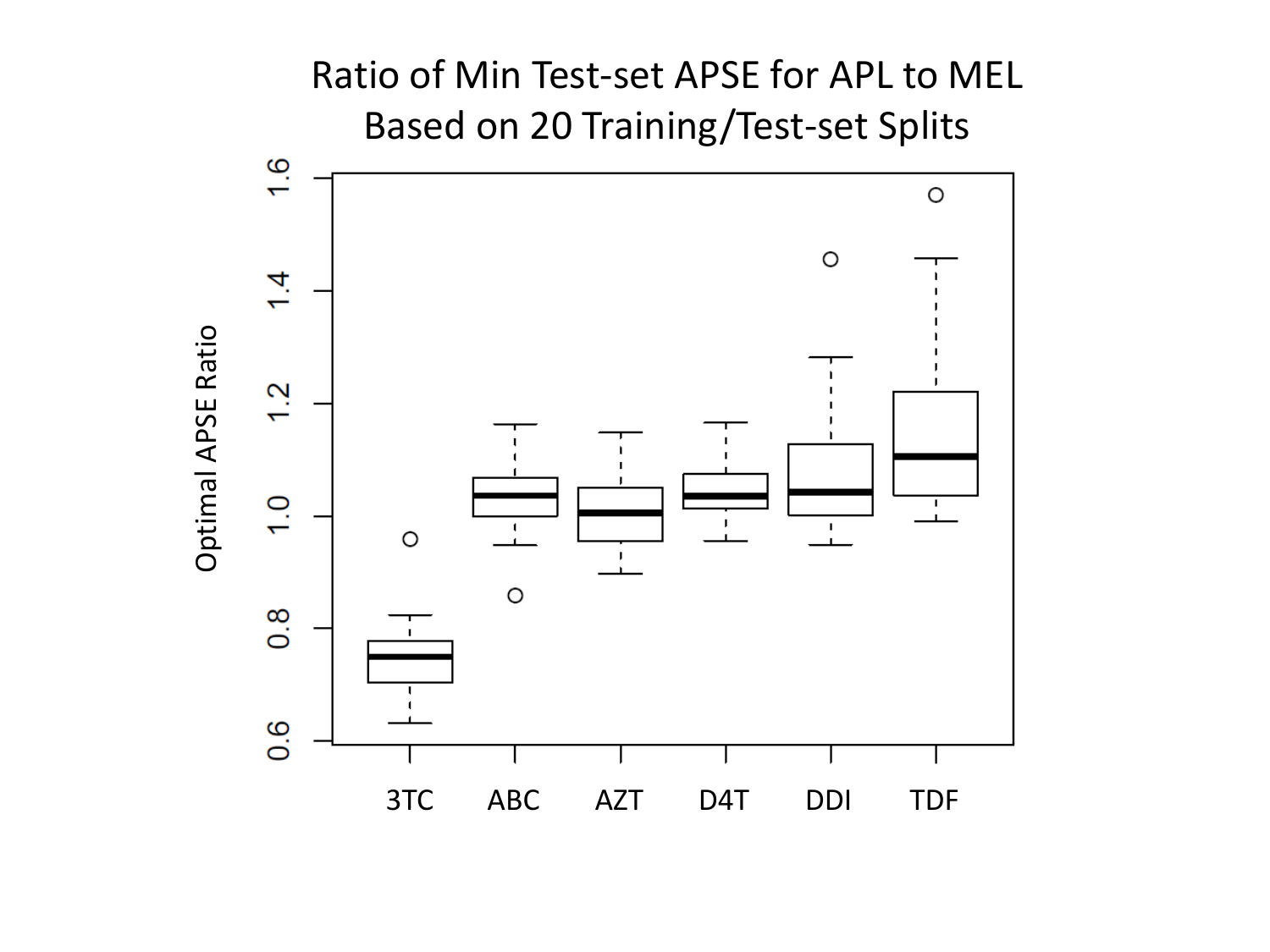}}
\caption{The ratio of the minimum test-set APSE obtained using the APL to that obtained using the MEL based on 20 random splits of the data for each of the six drugs.}
\label{f:ratios_HIV}
\end{figure}

Figure~\ref{f:ratios_HIV} plots the ratios of the minimum test-set APSE obtained using the APL to that obtained using the MEL based on 20 random splits of the data for each of the six drugs.

For the `3TC' drug, the inclusion of all pairwise interactions results in a dramatic improvement in performance.  In particular, there are five interactions that are included in all twenty of the selected models: `p62:p69',`p65:p184', `p67:p184', `p184:p215', and `p184:p190'.  This suggests that there is a strong interaction effect in this example, and that the interactions between these molecular targets are useful for the predicting drug susceptibility.

On the other hand, in four of the five remaining drugs - `ABC', `D4T', `DDI', and `TDF' - the inclusion of all pairwise interactions results in a significant deterioration in performance.  Here significance is determined using a Wilcoxon signed-rank test performed at a 0.05 significance level.  Thus, although the MEL is a restricted version of the APL, we still observe deterioration in the best-case predictive performance.  This suggests that although the Lasso allows the modeling of higher-order interactions, their inclusion should be done with care as doing so can hurt overall performance.

\section{Discussion}\label{S:concl}

The Lasso allows the fitting of regression models with a large number of predictor variables, but the resulting cost can be much higher than the loss bounds in the literature would suggest.  We have proven that when tuned optimally for prediction the performance of the Lasso deteriorates as the number of predictor variables increases with probability arbitrarily close to one
under the assumptions of a sparse true model with one true predictor and an orthonormal deterministic design matrix.  Our empirical results suggest that this deterioration persists as the sample size increases, and carries over to more general contexts.  

\begin{figure}[H]
\centering
\caption{Percentiles of the optimal loss ratios for SCAD over 1000 realizations as a function of $p$ for the orthogonal predictors example with $n=100$ and $p_0=6$.  The ``High SNR'' and ``Low SNR'' settings correspond to $\sigma^2=4$ and $\sigma^2=400$, respectively.}
\label{f:scad}
\subfigure[High SNR]{
\resizebox{6.5cm}{!}{\includegraphics{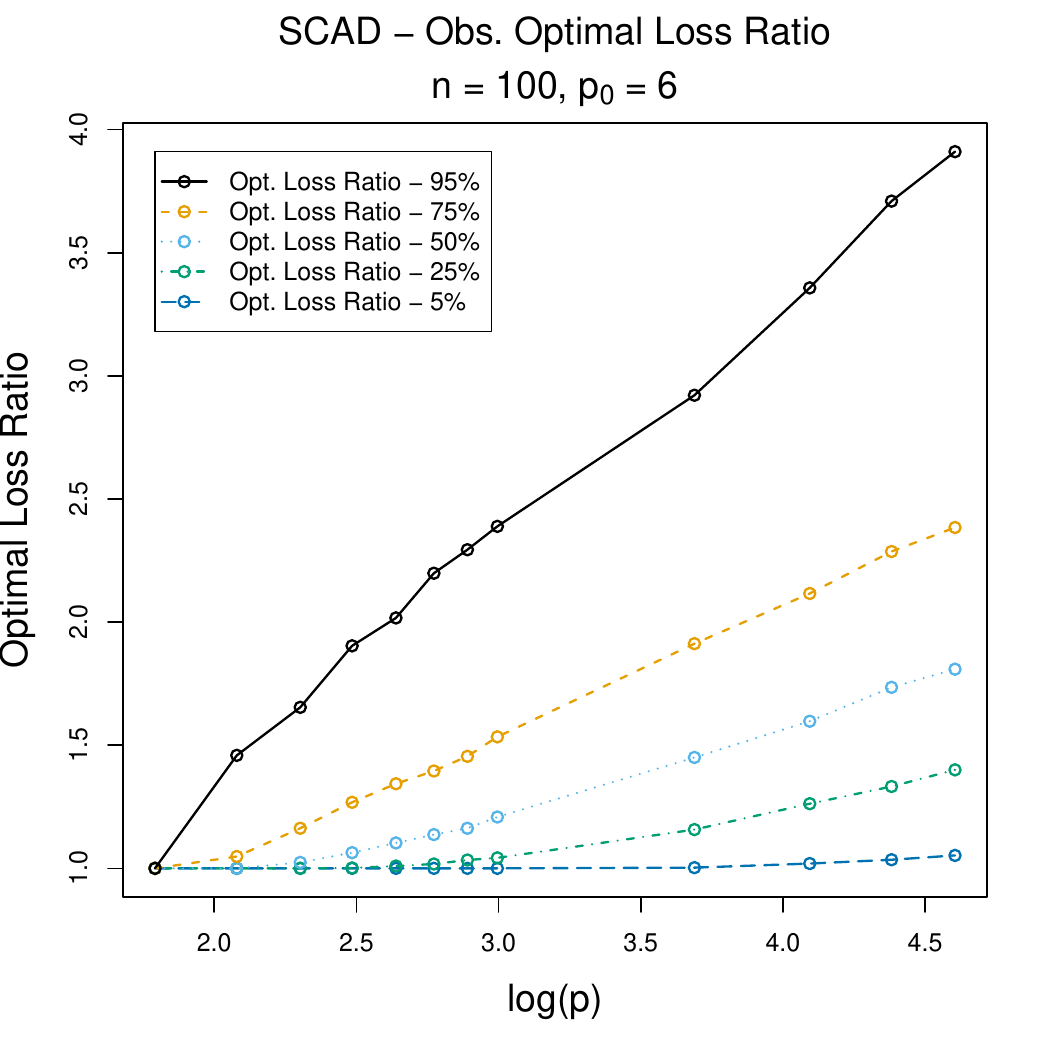}}}
\subfigure[Low SNR]{
\resizebox{6.5cm}{!}{\includegraphics{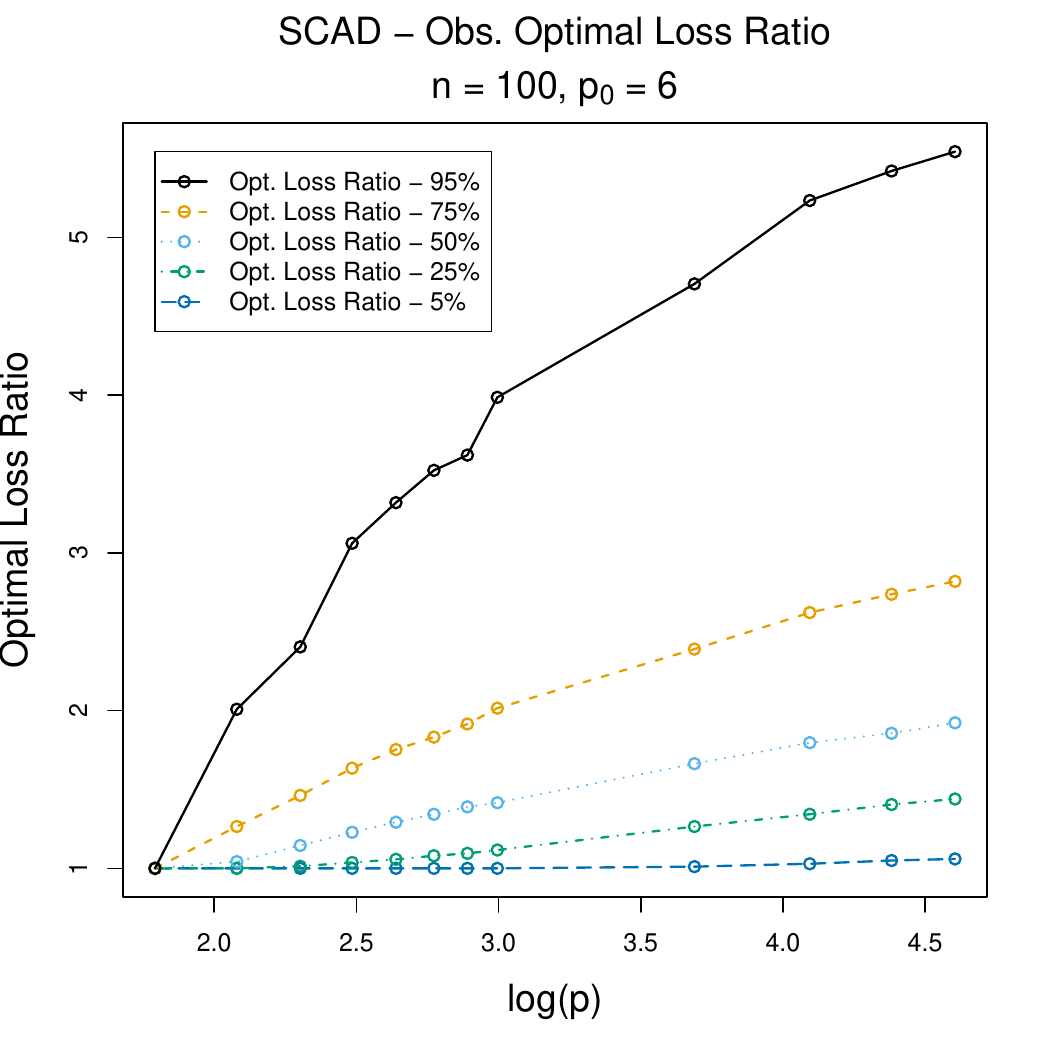}}}
\end{figure}

In classical all-subsets regression, deterioration in the optimal loss does not occur, because it is always possible to recover the estimated true model while ignoring the extraneous predictors.  This is not possible with the Lasso, because the only way to exclude extraneous predictors is to increase the amount of regularization imposed on all the estimated coefficients.  This property is not unique to the Lasso, and preliminary results suggest that deterioration also occurs when using other regularization procedures.  For example, Figure~\ref{f:scad} plots the percentiles of the optimal loss ratios for SCAD \citep{fan01} under the set-up of Section~\ref{S:orthog-pred} with orthogonal predictors.  In both plots there is evidence of deterioration.  However, comparing Figure~\ref{f:scad} to Figure~\ref{f:ratios_nfixed}, the degree of deterioration is typically less severe for SCAD than for the Lasso, especially in the High SNR setting.  This partly due to the fact that the SCAD penalty imposes less shrinkage on the estimated coefficients.  In the context of categorical predictors, \citet{flynn16} also found evidence of deterioration when working with the group Lasso and the ordinal group Lasso.  However, since the group Lasso and the ordinal group Lasso both impose more structure on the estimated coefficients, they reduce the effective degrees of freedom and the resulting observed deterioration for both methods is typically less severe than the deterioration observed when using the ordinary Lasso.  

\begin{figure}[H]
\centering
\subfigure[High SNR]{
\resizebox{6.5cm}{!}{\includegraphics[trim={2cm 1.5cm 2cm 2cm}]{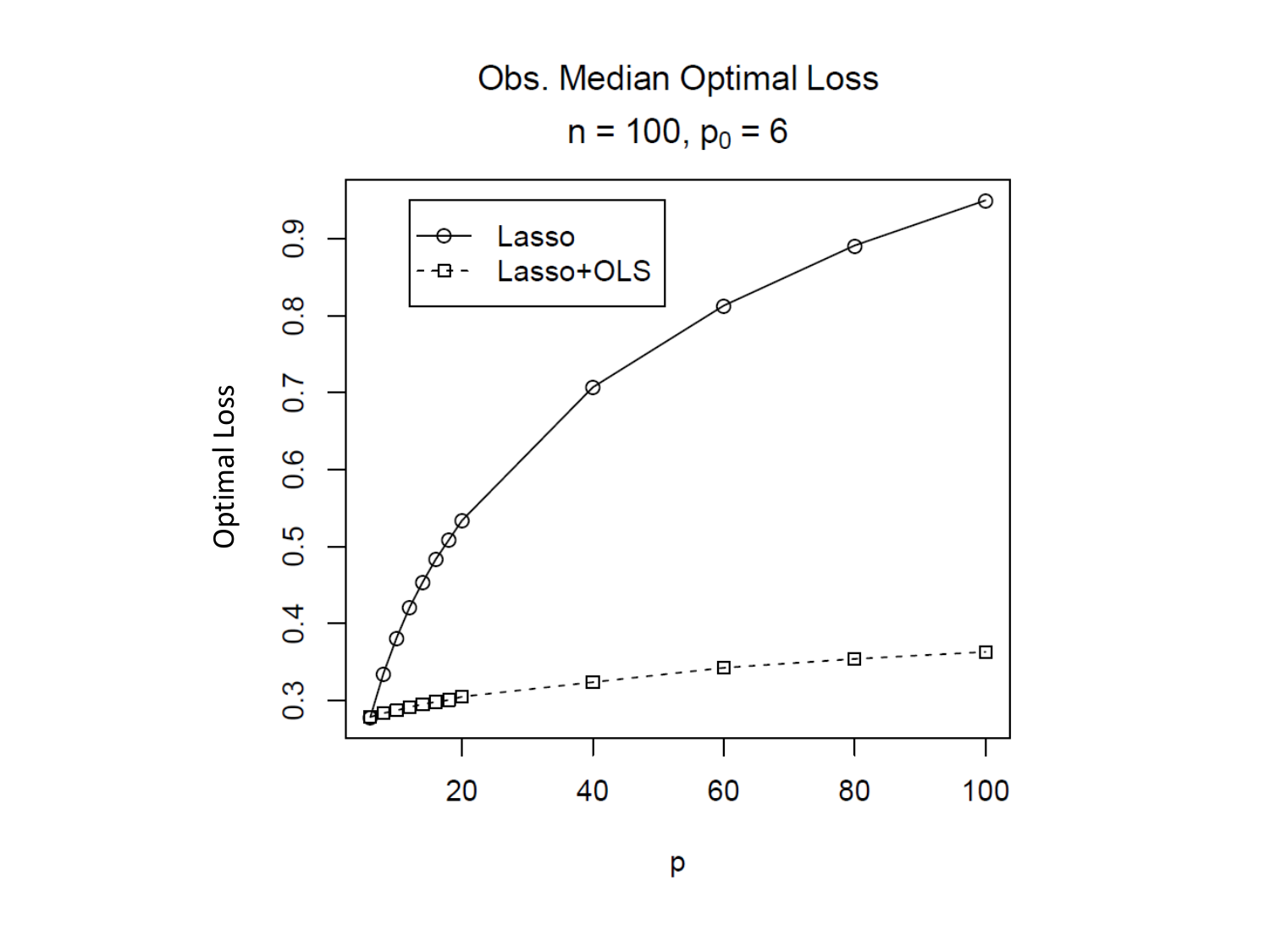}}}
\subfigure[Low SNR]{
\resizebox{6.5cm}{!}{\includegraphics[trim={2cm 1.5cm 2cm 2cm}]{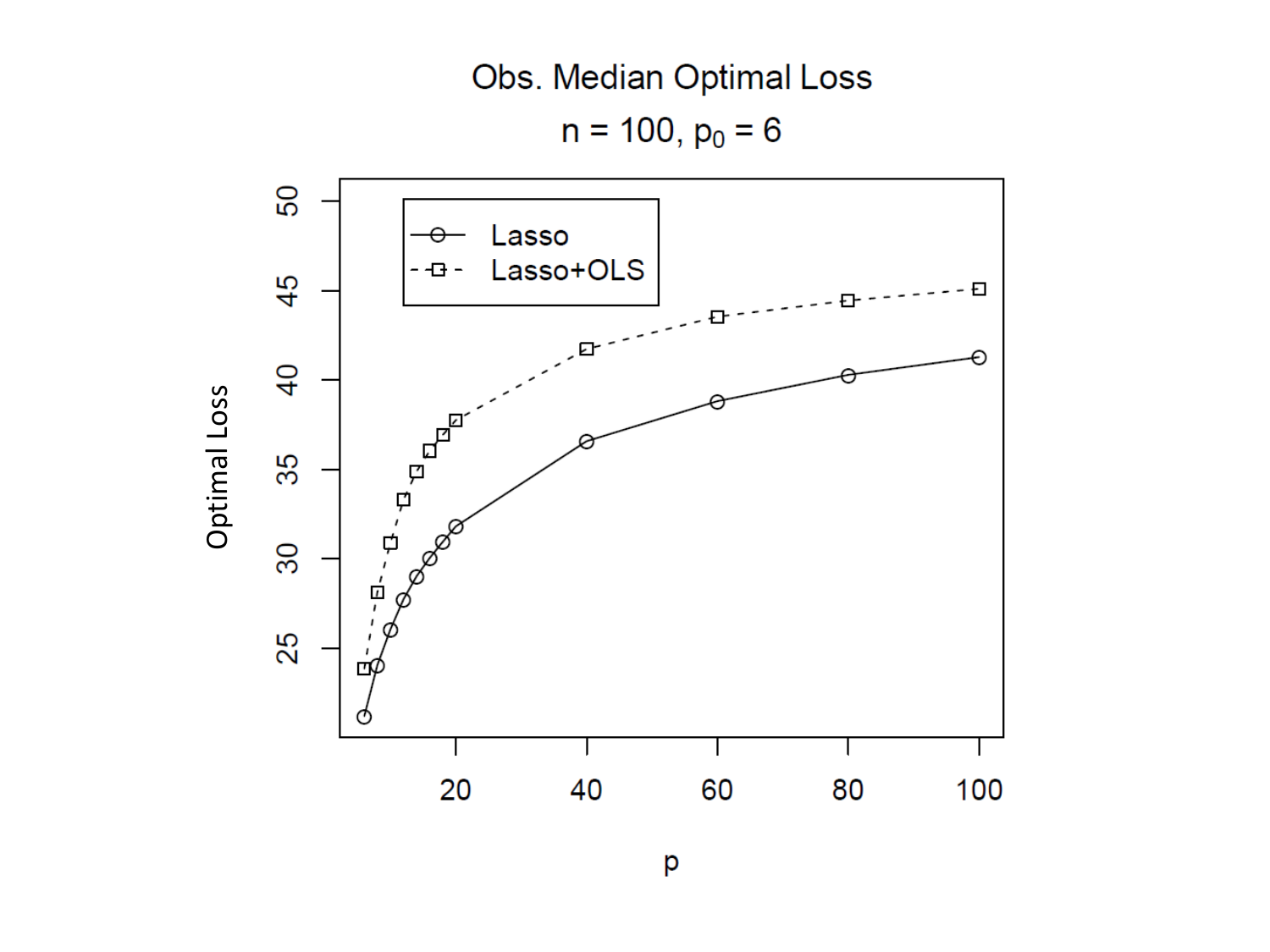}}}
\caption{Median optimal loss for the Lasso and Lasso+OLS over 1000 realizations as a function of $p$ for the orthogonal predictors example with $n=100$ and $p_0=6$.  The ``High SNR'' and ``Low SNR'' settings correspond to $\sigma^2=4$ and $\sigma^2=400$, respectively.}
\label{f:lasso-ols}
\end{figure}

In light of the deterioration in performance, data analysts should be careful when using the Lasso and other regularization procedures as variable selection and estimation tools with high-dimensional data sets.  One possible modification is to use the regularization procedure as a subset selector, but not as an estimation procedure.  One implementation of this is the extreme version of the Relaxed Lasso \citep{meinshausen07}, which fits least squares regressions to the Lasso selected subsets. Returning to the orthogonal predictors example in Section~\ref{S:orthog-pred}, we investigate the performance of this simple two-step procedure.  Figure~\ref{f:lasso-ols} plots the median optimal loss for the Lasso and the median optimal loss for the two-stage procedure for varying values of $p$.  In this example, the two-stage procedure improves performance when the SNR is high, but not when the SNR is low.  However, the improvement in performance in the high SNR case is more than the worsening of performance in the low SNR case.  These preliminary results suggest that a two-stage procedure that imposes no shrinkage on the estimated coefficients can help improve performance when the SNR is sufficiently high.  

Another possible solution is to screen the predictor variables before fitting the Lasso penalized regression.  In screening, the typical goal is to reduce from a huge scale to something that is $o(n)$ (Fan and Lv, 2008).  However, our results suggest that it is not enough to merely reduce the number of predictors, which implies that how to optimally tune the number of screened predictors is an interesting model selection problem.

One may also consider alternatives to regularization.  For example, \citet{ando14} achieved good performance in high-dimensional regression problems using a simple model averaging technique.  More recently, \citet{bertsimas2016} developed a Mixed Integer Optimization approach to best subset selection, which they found could outperform the Lasso in numerical experiments.  Further investigation into all of these techniques is an interesting area for future research.

\appendix

\section{Technical Results}\label{S:tech}

In this appendix we provide the proofs for the theoretical results presented in Section~\ref{S:theory}.

First we prove the results for the more general $p_0$-sparse case.
 
\begin{proof}[Lemma~\ref{l:s-sparse}]
First note that $nL_p(\lambda_p^*) \leq \sum_{j = 1}^{p_0} \beta^2_j$, because for any $\lambda_p^* \geq \max_{1 \leq j \leq p} |z_j|$, $nL_p(\lambda_p^*) = \sum_{j = 1}^{p_0} \beta^2_j$.  If $\lambda^*_{p_0} \geq \max_{1 \leq j \leq p_0} |z_j|$ then $nL_{p_0}(\lambda_{p_0}^*) = \sum_{j = 1}^{p_0} \beta^2_j$ and the
optimal $\lambda$ will be one such that all of the estimated coefficients equal zero.  No deterioration will occur in this case.

For the remainder of the proof assume that $\lambda^*_{p_0} \leq \max_{1 \leq j \leq p_0} |z_j|$.  Consider
\[
nL_p(\lambda_p^*) = nL_{p_0}(\lambda^*_p) + \sum_{j = p_0+1}^{p}(z_j - \lambda_p^*)^2_+ \geq nL_{p_0}(\lambda_p^*) \geq nL_{p_0}(\lambda_{p_0}^*).
\]
The optimal loss does not deteriorate when equality holds.

If $\lambda^*_{p_0} \geq \max_{p_0 < j \leq p} |z_j|$, then
\[
nL_{p_0}(\lambda_{p_0}^*) = nL_{p_0}(\lambda_{p_0}^*) + \sum_{j = p_0+1}^{p}(z_j - \lambda_p^*)^2_+ = L_p(\lambda_{p_0}^*).
\]
This implies that $nL_p(\lambda^*_p) = nL_{p_0}(\lambda^*_{p_0})$ and no deterioration occurs.

Alternatively, if $\lambda^*_{p_0} < \max_{p_0 < j \leq p} |z_j|$, then
\[
L_s(\lambda^*_{p_0}) < L_{p_0}(\lambda^*_{p_0}) + \sum_{j=p_0+1}^{p}(z_j - \lambda_{p_0}^*)^2_+ = L_p(\lambda^*_{p_0}) \leq L_p(\lambda_p^*),
\]
so the optimal loss deteriorates.

It follows that deterioration occurs if and only if $\lambda^*_{p_0} \leq \max_{1 \leq j \leq p_0} |z_j|$ and $\lambda^*_{p_0} < \max_{p_0 < j \leq p} |z_j|$.
\end{proof}

\begin{proof}[Theorem~\ref{T:prob-det-s-sparse}]
By Lemma~\ref{l:s-sparse},
\begin{align*}
\Pr \left( \frac{L_{p}(\lambda_p^*)}{L_{p_0}(\lambda_{p_0}^*)} > 1 \right)
	& = \Pr(\lambda^*_{p_0} \leq \max_{1 \leq j \leq p_0} |z_j| \, ,
			\lambda^*_{p_0} \leq \max_{p_0 < j \leq p} |z_j|) \\
	& \geq \Pr(0 \leq \max_{1 \leq j \leq p_0} |z_j| \, ,
				0 \leq \max_{p_0 < j \leq p} |z_j| \, ,
				\lambda^*_{p_0} = 0) \\
	& = \Pr(\lambda^*_{p_0} = 0).
\end{align*}
Therefore, it is sufficient to show that
\(
\Pr(\lambda^*_{p_0} = 0) > 0
\)
to show that the probability of deterioration is non-zero.

Consider the set
\[
\sS \equiv \{z_j, 1 \leq j \leq p_0 : \beta_1 > z_1 > \beta_2 > z_2 > \ldots > \beta_{p_0} > z_{p_0} > 0 \}.
\]
Assume that $z_j, 1 \leq j \leq p_0 \in \sS$.  This implies that
\(
\sum_{j = 1}^k (\beta_j - z_j) > 0
\)
for all $1 \leq k \leq p_0$.

For any $\lambda \in [0, z_{p_0})$,
\[
nL_{p_0}(\lambda) = \sum_{j=1}^{p_0} (\beta_j - (z_j - \lambda))^2,
\]
and
\[
\frac{\partial nL_{p_0}(\lambda)}{\partial \lambda}
	= 2 \sum_{j=1}^{p_0} (\beta_j - z_j) + 2 p_0 \lambda.
\]
Since the derivative is an increasing function of $\lambda$ and it is non-negative at $\lambda = 0$, the minimum occurs at $\lambda = 0$.

Next, for any $1 < k \leq p_0$, consider $\lambda \in I_k = (z_k, z_{k-1})$.  Over this interval,
\[
nL_{p_0}(\lambda) = \sum_{j=1}^{k-1} (\beta_j - (z_j - \lambda))^2 + \sum_{j=k}^{p_0} \beta_j^2,
\]
and
\[
\frac{\partial nL_{p_0}(\lambda)}{\partial \lambda}
	= 2 \sum_{j=1}^{k-1} (\beta_j - z_j) + 2 (k-1) \lambda.
\]

Since the derivative is an increasing function of $\lambda$ and it is non-negative at $\lambda = z_k$, the minimum occurs at $\lambda = z_k$.  However, for any $1 < k \leq p_0$,
\[
nL_{p_0}(z_k) = \sum_{j=1}^{k-1} (\beta_j - (z_j - z_k))^2 + \sum_{j=k}^{p_0} \beta_j^2
 > \sum_{j=1}^{p_0} (\beta - z_j))^2 = nL_{p_0}(0).
\]
Thus, $\lambda_{p_0}^* \not\in I_k$ for any $1 < k \leq p_0$.

Finally, for any $\lambda \in [z_1, \infty)$,
\[
nL_{p_0}(\lambda) = \sum_{j=1}^{p_0} \beta_j^2 > \sum_{j=1}^{p_0}(\beta_j - z_j)^2 = nL_{p_0}(\lambda).
\]
It follows that $\lambda_{p_0}^* = 0$ on $\sS$.

Since the $z_j$'s, $1 \leq j \leq p_0$ are independent normal random variables, it follows that
\[
\Pr(\lambda_{p_0}^* = 0) \geq \Pr(\sS) > 0.
\]
Thus, equation~(\ref{e:det-lower-bound}) is satisfied.
\end{proof}

Next, to prove Theorem~\ref{T:prob-det}, we establish the following four lemmas.  
First note that one can always choose $\lambda \geq \max_{1\leq j \leq p} |z_j|$,
which will shrink all of the estimated coefficients to zero.
Thus, for all $p>0$, $nL_p(\lambda^*_p) \leq \beta_1^2$.
The following Lemma establishes that equality always occurs if the sign of $z_1$ is incorrect.

\begin{lemma}\label{l:worst-case}
If $\sgn(\beta_1) \neq \sgn(z_1)$, then $nL_p(\lambda^*_p) = \beta_1^2$ for all $0<p\leq n$.
\end{lemma}

\begin{proof}
If $\sgn(\beta_1) \neq \sgn(z_1)$, then for any $\lambda < \max_{1\leq j \leq p} |z_j|$,
\[
nL_p(\lambda)
= (\beta_1 - \sgn(z_1)(|z_1| - \lambda)_+)^2 + \sum_{j=2}^p (|z_j|-\lambda)_+^2
\geq \beta_1^2 + \sum_{j=2}^p (|z_j|-\lambda)_+^2
\geq \beta_1^2.
\]
Thus $nL_p(\lambda_p^*) = \beta_1^2$.
\end{proof}

Lemma~\ref{l:worst-case} establishes that if the sign of $z_1$ is incorrect, $L_p(\lambda_p^*)=L_1(\lambda_1^*)$ for all $p>1$, so no deterioration will occur.

Next we focus our attention on the situation where the sign of $z_1$ is correct.  The following lemma establishes the
optimal loss for the Lasso when only the one true predictor is used.

\begin{lemma}\label{l:true-pred-loss}
If $\sgn(\beta_1)=\sgn(z_1)$, then
\[ nL_1(\lambda_1^*) = \left\{
  \begin{array}{l l}
    0 & \quad \text{if $|\beta_1| \leq |z_1|$}\\
    (\beta_1-z_1)^2 & \quad \text{otherwise}
  \end{array} \right..\]
\end{lemma}

\begin{proof}
Without loss of generality assume that $\beta_1 >0$, and therefore $z_1>0$.  Consider
\[
nL_1(\lambda) = (\beta_1 - (z_1 - \lambda)_+)^2.
\]
First consider $\lambda \in I = [0, z_1)$.  Since $nL_1(\lambda)$ is a convex function for $\lambda \in I$, the minimum
occurs at a place where the
derivative is zero or when $\lambda=0$.
Taking the derivative with respect to $\lambda \in I$,
\[
\frac{ \partial nL_1(\lambda)}{\partial \lambda}=
    2(\beta_1 - (z_1 - \lambda)) .
\]
Since the derivative is an increasing function of $\lambda$, a minimum occurs at $\lambda=0$ if the
derivative is non-negative at that point.  In other words, a minimum occurs at $\lambda = 0 $ if $\beta_1 \geq z_1$.  Otherwise, a
minimum occurs at a point where the derivative is zero.
Thus
\[
\argmin_{\lambda \in I} nL_1(\lambda) = \left\{
  \begin{array}{l l}
    z_1 - \beta_1 & \quad \text{if $0 \leq \beta_1 < z_1 $}\\
    0 & \quad \text{if $z_1 \leq  \beta_1 $}
  \end{array} \right.,\]
and
\[ \min_{\lambda \in I} nL_1(\lambda) = \left\{
  \begin{array}{l l}
    0 & \quad \text{if $0 \leq \beta_1 < z_1 $}\\
    (\beta_1-z_1)^2 & \quad \text{if $z_1 \leq  \beta_1 $}
  \end{array} \right..\]
Next, for $\lambda \geq z_1$, $nL(\lambda) = \beta_1^2$.  Since $\min_{\lambda \in I} nL(\lambda) < \beta_1^2$ for all $\beta_1>0$,
it follows that
\[ nL_1(\lambda_1^*) = \left\{
  \begin{array}{l l}
    0 & \quad \text{if $0 \leq \beta_1 < z_1 $}\\
    (\beta_1-z_1)^2 & \quad \text{if $z_1 \leq  \beta_1 $}
  \end{array} \right..\]
\end{proof}

In this case, when the model includes superfluous predictors, the optimal level of shrinkage is determined by
balancing the increase in loss due to the bias induced from over-shrinking
the true estimated coefficient
with the increase in loss due to under-shrinking the estimated coefficients for the superfluous predictors.
The next two lemmas establish necessary and sufficient conditions on the $z_j$'s for deterioration to occur.

\begin{lemma}\label{l:no-det-case}
Assume that $\sgn(\beta_1)=\sgn(z_1)$.  If $\max_{2 \leq j \leq p} |z_j| < |z_1|$, then $L_p(\lambda_p^*)=L_1(\lambda_1^*)$
if and only if $|\beta_1|<|z_1| - \max_{2 \leq j \leq p} |z_j|$.
\end{lemma}

\begin{proof}
Without loss of generality assume that $\beta_1 > 0$, and therefore $z_1>0$.  Also assume that
$|z_2|>\ldots>|z_p|$.
Consider
\[
nL_p(\lambda) = (\beta_1 - (z_1 - \lambda)_+)^2 + \sum_{j=2}^p (|z_j| - \lambda)_+^2.
\]

First consider $\lambda \in I = [0,z_1)$.  Since $nL_p(\lambda)$ is a continuous differentiable function for $\lambda \in I$, local extrema
occur at points where the derivative is zero or at a boundary point.  Taking the derivative with respect to $\lambda$,
\[
\frac{ \partial nL_p(\lambda)}{\partial \lambda}= \left\{
  \begin{array}{l l}
    2(\beta_1 - (z_1 - \lambda)) & \quad \text{if $|z_2| \leq \lambda < z_1 $}\\
    2(\beta_1 - (z_1 - \lambda)) - 2 \sum_{j=2}^k (|z_j|-\lambda)
        & \quad \begin{tabular}{@{}l@{}} \text{if $|z_{k+1}| \leq  \lambda < |z_k|, $ }\\
                    \quad \text{for $k=2,\ldots,p-1$}\end{tabular}\\
    2(\beta_1 - (z_1 - \lambda)) - 2 \sum_{j=2}^p (|z_j|-\lambda)
        & \quad \text{if $0 \leq  \lambda < |z_p|$}\\
  \end{array} \right..\]
Since the derivative is a strictly increasing function of $\lambda$, a minimum occurs at $\lambda=0$ if the
derivative is non-negative at that point.  Hence a minimum occurs at $\lambda = 0 $ if $\beta_1 > \sum_{j=1}^p |z_j|$.
Otherwise a minimum occurs at a point where the derivative is zero.  Define
\[
\lambda_{I}^* \equiv \argmin_{\lambda \in I} nL_p(\lambda).
\]
It follows that
\[
\lambda_{I}^* = \left\{
  \begin{array}{l l}
    z_1 - \beta_1 & \quad \text{if $0 < \beta_1 \leq z_1 - |z_2|$}\\
    \frac{\sum_{j=1}^k |z_j| - \beta_1}{k}
        & \quad \begin{tabular}{@{}l@{}} \text{if $\sum_{j=1}^k |z_j| - k|z_k| < \beta_1 \leq \sum_{j=1}^{k}|z_j| - k|z_{k+1}|$, }\\
                   \quad \text{for $k = 2, \ldots, p-1$} \end{tabular}\\
    \frac{\sum_{j=1}^p |z_j| - \beta_1}{p}
        & \quad \text{if $\sum_{j=1}^p |z_j| - p|z_p| < \beta_1 \leq \sum_{j=1}^{p}|z_j| $}\\
    0 & \quad \text{if $\sum_{j=1}^p |z_j| <  \beta_1 $}
      \end{array} \right..\]

Next, for $\lambda \geq z_1$, $nL_p(\lambda) = \beta_1^2$.  Thus
\[
nL_p(\lambda^*_p) = \min\Big(\beta_1^2, nL_p(\lambda^*_I)\Big).
\]

To compare $nL_p(\lambda_p^*)$ to $nL_1(\lambda_1^*)$, first note that $nL_1(\lambda_1^*) < \beta_1^2$.
Next, comparing $nL_p(\lambda_I^*)$ to $nL_1(\lambda_1^*)$ it is clear that $nL_1(\lambda_1^*) = nL_p(\lambda_I^*)=0$
if $0<\beta_1 \leq z_1-|z_2|$.
However, if $z_1-|z_2|<\beta_1 \leq z_1$,
then $\lambda_I^* < |z_2|$ and
\[
nL_p(\lambda_I^*) > (|z_2|-\lambda_p^*)^2 > 0 = nL_1(\lambda_1^*).
\]
Similarly, if $z_1 < \beta_1$, then either $\lambda_I^* > 0$ so that
\[
nL_p(\lambda_I^*) > (\beta_1-(z_1-\lambda_I^*))^2 > (\beta_1 - z_1)^2 = nL_1(\lambda_1^*),
\]
or $\lambda_I^*=0$ and
\[
nL_p(\lambda_I^*) = (\beta_1-z_1)^2 + \sum_{j=2}^p|z_j|^2 > (\beta_1 - z_1)^2 = nL_1(\lambda_1^*).
\]
Hence, $nL_1(\lambda_1^*) = nL_p(\lambda_p^*)$ if and only if $0<\beta_1 \leq z_1-|z_2|$.
\end{proof}

\begin{lemma}\label{l:always-det}
Assume that $\sgn(\beta_1)=\sgn(z_1)$.  If $\max_{2 \leq j \leq p} |z_j| > |z_1|$, then $L_p(\lambda_p^*)>L_1(\lambda_1^*)$
for all $\beta_1 \neq 0$.
\end{lemma}

\begin{proof}
Without loss of generality assume that $\beta_1 > 0$, and therefore $z_1>0$.  Also assume that
$|z_2|>\ldots>|z_p|$.
Consider
\[
nL_p(\lambda) = (\beta_1 - (z_1 - \lambda))^2 + \sum_{j=2}^p (|z_j| - \lambda)_+^2.
\]
Define
\[
\tilde{k} = \max_{2 < k \leq p} \{k:|z_k|>z_1\}.
\]
The derivative of $nL_p(\lambda)$ does not exist at $\lambda = z_1$.
However, by Lemma~\ref{l:worst-case}, $\lambda=z_1$ is never globally optimal since
\[
nL_p(z_1) = \beta_1^2 + \sum_{j=2}^{\tilde{k}} (|z_j|-z_1)^2 > \beta_1^2.
\]
To determine the optimal values of $\lambda$, we consider the intervals $I_1 = [0,z_1)$,
$I_2 = (z_1, |z_2|]$, and $I_3 = (|z_2|, \infty)$ separately.  Define
\[
\lambda^*_{I_j} = \argmin_{\lambda \in I_j} nL_p(\lambda)
\]
for $j=1,2,3$.

First, for $\lambda \in I_1$, $nL_p(\lambda)$ is a continuous differentiable function
and
\[
\frac{ \partial nL_p(\lambda)}{\partial \lambda}= 2(\beta_1 - (z_1 - \lambda)) - 2\sum_{j=2}^p (|z_j|-\lambda)_+.
\]
Since the derivative is a strictly increasing function of $\lambda$, a minimum occurs at $\lambda=0$ if the
derivative is non-negative at that point.  Thus, a minimum occurs at $\lambda = 0 $ if $\beta_1 > \sum_{j=1}^p |z_j|$.
Similarly, a local minimum occurs at $\lambda=z_1$ if
\[
\lim_{\lambda \to z_1^{-}} \frac{ \partial nL_p(\lambda)}{\partial \lambda} < 0,
\]
which holds if $0< \beta_1 \leq \sum_{j=1}^{\tilde{k}} |z_j| - \tilde{k}z_1$.
Otherwise, a
minimum occurs at a point where the derivative is zero.  It follows that
\[
\lambda_{I_1}^* = \left\{
  \begin{array}{c l}
    z_1
    & \quad \text{if
            $0< \beta_1 \leq \sum_{j=1}^{\tilde{k}} |z_j| - \tilde{k}z_1$} \\
    \frac{\sum_{j=1}^{\tilde{k}} |z_j| - \beta_1}{\tilde{k}}
        & \quad \text{if
            $\sum_{j=1}^{\tilde{k}} |z_j| - \tilde{k}z_1
            < \beta_1 \leq
            \sum_{j=1}^{\tilde{k}} |z_j| - \tilde{k}|z_{\tilde{k}+1}|$, }\\
    \frac{\sum_{j=1}^k |z_j| - \beta_1}{k}
       & \quad \begin{tabular}{@{}l@{}} \text{if
            $\sum_{j=1}^{k} |z_j| - k|z_k|
            < \beta_1 \leq
            \sum_{j=1}^k |z_j| - k|z_{k+1}|$, }\\
        \quad \text{for $k=\tilde{k}+1,\ldots,p-1$}\end{tabular}\\
    \frac{\sum_{j=1}^p |z_j| - \beta_1}{p}
        & \quad \text{if $\sum_{j=1}^p |z_j| - p|z_p| < \beta_1 \leq \sum_{j=1}^{p}|z_j| $}\\
    0 & \quad \text{if $\sum_{j=1}^p |z_j| <  \beta_1 $}
      \end{array} \right..\]

Next, for $\lambda \in I_2$, $nL_p(\lambda)$ is a continuous differentiable function and
\[
\frac{ \partial nL_p(\lambda)}{\partial \lambda}= - 2\sum_{j=2}^p (|z_j|-\lambda)_+.
\]
Since the derivative is negative for all $\lambda \in I_2$, a local minimum occurs at $\lambda = |z_2|$, thus
$nL_p(\lambda^*_{I_2}) = \beta_1^2$.

Lastly, for all $\lambda \in I_3$, $nL_p(\lambda) = \beta_1^2$.  It follows that

\[
nL_p(\lambda_p^*) = \min\Big( \beta_1^2, nL_p(\lambda^*_{I_1}) \Big)
\]
By a similar argument to that used in the proof of Lemma~\ref{l:no-det-case}, it follows that $nL_p(\lambda_p^*) > nL_1(\lambda_1^*)$.
\end{proof}

It follows that deterioration occurs unless it is possible to shrink $z_1$ optimally while at the same time shrinking all of the estimated coefficients for the superfluous predictors to zero.  In particular,
by Lemmas~\ref{l:no-det-case} and \ref{l:always-det}, when the sign of $z_1$ is correct, $L_p(\lambda_p^*)>L_1(\lambda_1^*)$ unless $|\beta_1|<|z_1| - \max_{2 \leq j \leq p} |z_j|$.

\begin{proof}[Theorem~\ref{T:prob-det}]

By Lemma~\ref{l:worst-case},
\begin{multline*}
\Pr \big( nL_p(\lambda_p^*) > nL_1(\lambda_1^*) \big) \\
= \Pr \big( nL_p(\lambda_p^*) > nL_1(\lambda_1^*) \big\vert \sgn(z_1)=\sgn(\beta_1)\big) \Pr \big(\sgn(z_1)=\sgn(\beta_1)\big).
\end{multline*}
Without loss of generality, assume that $\beta_1>0$.
By Lemmas~\ref{l:no-det-case}-\ref{l:always-det}, this is equal to
\begin{multline*}
\Big(1-\Pr \big( nL_p(\lambda_p^*) = nL_1(\lambda_1^*) \big\vert z_1>0\big)\Big) \Pr\big(z_1>0\big)\\
=\Big(1-\Pr \big( \beta_1 < z_1 - \max_{2 \leq j \leq p} |z_j| \big\vert z_1>0\big)\Big)\Pr\big(z_1>0\big).
\end{multline*}
We can evaluate these probabilities explicitly.  First consider
\begin{equation}\label{e:prob-p1}
\Pr( z_1 > 0 ) = \Phi \left(\frac{\beta_1}{\sigma}\right).
\end{equation}
Next,
\begin{align*}
\Pr \big( \beta_1 < z_1 - \max_{2 \leq j \leq p} |z_j| \big\vert z_1>0\big)
& =
\frac{ \Pr \big( \{\beta_1 < z_1 - \max_{2 \leq j \leq p}|z_j| \} \cap \{z_1>0\} \big) }
{\Pr \big( z_1>0 \big)}\\
&= \frac{ \Pr \big( \beta_1 < z_1 - \max_{2 \leq j \leq p} |z_j| \big) }
{\Pr \big( z_1>0 \big)},
\end{align*}
where the second equality follows from the fact that $\beta_1>0$ implies that $z_1>0$.
By~(\ref{e:z1-dist}) and (\ref{e:zj-dist}),
\begin{align*}
\Pr \big( \beta_1 < z_1 - \max_{2 \leq j \leq p} |z_j| \big)
& = \Pr \big( \cap_{j=2}^p \{\beta_1 < z_1 - |z_j| \} \big) \\
& = \int_{z_1=\beta_1}^{z_1=\infty} \left[ \int_{z_2=-(z_1-\beta_1)}^{z_2=z_1-\beta_1}
    f_2(z_2)\mathrm{d}z_2 \right]^{p-1} f_1(z_1) \mathrm{d}z_1\\
& = \int_{\beta_1}^{\infty} \left[ 2\Phi \left( \frac{z_1 - \beta_1}{\sigma} \right) -1 \right]^{p-1} f_1(z_1)
\mathrm{d}z_1 \\
& = \frac{1}{\sigma} \int_{\beta_1}^\infty
        \left[ 2\Phi \left( \frac{z_1 - \beta_1}{\sigma} \right) -1 \right]^{p-1}
        \phi\left(\frac{z_1-\beta_1}{\sigma}\right) \mathrm{d}z_1,
\end{align*}
where $f_1(\cdot)$ and $f_2(\cdot)$ are the probability distribution functions (pdf) of $z_1$ and $z_2$, respectively,
and $\phi(\cdot)$ is the pdf of the standard normal distribution.
Substituting
\[
w = 2\Phi \left( \frac{z_1 - \beta_1}{\sigma} \right) -1,
\]
\[
\Pr\big(  \beta_1 < z_1 - \max_{2 \leq j \leq p} |z_j|\big)
= \frac{1}{2} \int_0^1 w^{p-1} \mathrm{d}w = \frac{1}{2p}.
\]
Thus
\begin{equation}\label{e:prob-p2}
\Pr \big( \beta_1 < z_1 - \max_{2 \leq j \leq p} |z_j| \big\vert z_1>0 \big)
= \frac{ \frac{1}{2p}}{\Phi \Big(\frac{\beta_1}{\sigma}\Big)}.
\end{equation}
From (\ref{e:prob-p1}) and (\ref{e:prob-p2}), it follows that
\[
\Pr ( L_p(\lambda_p^*) > L_1(\lambda_1^*) ) = \Phi \left(\frac{\beta_1}{\sigma}\right) - \frac{1}{2p}.
\]

\end{proof}

Lastly, we provide the proof for Theorem~\ref{T:exp-det}]

\begin{proof}[Theorem~\ref{T:exp-det}]
Without loss of generality, assume that $\beta_1>0$.  Define
\[
\mathcal{A} := \{ z_1, z_2 : z_2 > z_1 > \beta_1 \}.
\]
Note that $\Pr \Big((z_1,z_2) \in \mathcal{A} \Big)>0$.  By Lemmas~\ref{l:true-pred-loss} and \ref{l:always-det},
for $(z_1,z_2) \in \mathcal{A}$, $L_1(\lambda_1^*)=0$ and $L_p(\lambda_p^*)> L_1(\lambda_1^*)$.  Thus,
\[
\frac{L_p(\lambda_p^*)}{L_1(\lambda_1^*)} = \infty
\]
It follows that
\begin{align*}
\E \left( \frac{L_p(\lambda_p^*)}{L_1(\lambda_1^*)} \right)
& = \int_{z_p=-\infty}^{z_p=\infty}\cdots\int_{z_1=-\infty}^{z_1=\infty}
\frac{L_p(\lambda_p^*)}{L_1(\lambda_1^*)}
df_1(z_1)\cdots df_p(z_p) \\
& \geq
\int_{z_p=-\infty}^{z_p=\infty}\cdots\iint\limits_{ (z_1, z_2) \in \mathcal{A}}
\frac{L_p(\lambda_p^*)}{L_1(\lambda_1^*)}
df_1(z_1)\cdots df_p(z_p) \\
& = \infty.
\end{align*}
\end{proof}

\section{Deterioration with two true predictors}\label{app:s-2det}
In this appendix we assume the same set-up as Section~\ref{S:theory}, where we assume that $p_0 = 2$.  Without loss of generality we assume that $\beta_1 > \beta_2 > 0$.

To compute the probability of deterioration, we first study the behavior of $\lambda_2^*$. 

\noindent \textbf{Case 1: $\mathbf{z_1, z_2 > 0}$}

For any $\lambda \geq 0$
\[
nL_s(\lambda) = \sum_{j = 1}^{2} (\beta_j - (z_j - \lambda)_+)^2.
\]

To simplify notation, define
\[
z^{(2)} = \max(z_1, z_2)
\]
and
\[
z^{(1)} = \min(z_1, z_2),
\]
and let $\beta^{(1)}$ and $\beta^{(2)}$ be the corresponding values of $\beta$, respectively.

We first compute the optimal value of $\lambda$ over a series of disjoint intervals: $I_1 = [0, z^{(1)})$, $I_2 = (z^{(1)}, z^{(2)})$ and $I_3 = [z^{(2)}, \infty)$.  Here we have excluded $\lambda = z^{(1)}$, but this point is never optimal.  Using similar techniques as those used in Lemmas~\ref{l:true-pred-loss}-\ref{l:always-det}, it follows that
\[
\argmin_{\lambda \in I_1} nL_2(\lambda) = \left\{
  \begin{array}{l l}
    z^{(1)} & \quad \text{if $0 \leq \beta^1 + \beta^2 \leq z^2 - z^1$}\\
    \sum_{j = 1}^{2}(z^{(j)} - \beta^{(j)})/2 & \quad \text{if $z^{(2)} - z^{(1)} < \beta^{(1)} + \beta^{(2)} \leq z^{(2)} + z^{(1)}$}\\
    0 & \quad \text{if $z^{(2)} + z^{(1)} < \beta^{(1)} + \beta^{(2)}$}
  \end{array} \right.,\]
\[
\argmin_{\lambda \in I_2} nL_2(\lambda) = \left\{
  \begin{array}{l l}
    z^{(2)} - \beta^{(2)} & \quad \text{if $0 \leq \beta^{(2)} \leq z^{(2)} - z^{(1)} $}\\
    z^{(1)} & \quad \text{if $z^{(2)} - z^{(1)} < \beta^{(2)}$}\\
  \end{array} \right.,\]
and for all $\lambda \in I_3$
\[
nL_2(\lambda) = \beta_1^{(2)} + \beta_2^{(2)},
\]
which is the worst-case loss.  By comparing the optimal loss for each interval, it can be shown that the global optimal choice for $\lambda$ is

\[
\lambda_2^* = \left\{
  \begin{array}{l l}
    z^{(2)} - \beta^{(2)} & \quad \text{if $0 \leq \beta^{(1)} + \beta^{(2)} \leq z^{(2)} - z^{(1)}$}\\
    \sum_{j = 1}^{2}(z^{(j)} - \beta^{(j)})/2 & \quad \text{if $z^{(2)} - z^{(1)} < \beta^{(1)} + \beta^{(2)} \leq z^{(2)} + z^{(1)}$}\\
    0 & \quad \text{if $z^{(2)} + z^{(1)} < \beta^{(1)} + \beta^{(2)}$}
  \end{array} \right..\]

\noindent \textbf{Case 2: $\mathbf{z_1, z_2 < 0}$}

In this case the signs of the estimated coefficients are incorrect.  Thus, for any $0 \leq \lambda < \max_{1 \leq j \leq 2} |z_j|$,
\[
nL_2(\lambda) = (\beta_1 + (|z_1| - \lambda)_+)^2 + (\beta_2 + (|z_2| - \lambda)_+)^2
	> \beta_1^2 + \beta_2^2.
\]
This implies that
\[
\lambda_2^* \geq \max_{1 \leq j \leq 2} |z_j|.
\]
Applying Theorem~\ref{l:s-sparse}, it follows that no deterioration occurs in this case.

\noindent \textbf{Case 3a: $\mathbf{z_1 < 0}$, $\mathbf{z_2 > 0}$, $\mathbf{|z_1| > |z_2|}$}

For any $\lambda \in [0, |z_2|)$
\begin{align*}\label{e:inc-signs-bound}
nL_2(\lambda) - nL_2(|z_1|) & = (\beta_1 + (|z_1| - \lambda)_+)^2 + (\beta_2 - (z_2 - \lambda)_+)^2  - (\beta_1^2 + \beta_2^2)\\
& = 2\beta_1(|z_1| - \lambda) - 2\beta_2(z_2 - \lambda) + (z_2 - \lambda)^2 + (|z_1| - \lambda)^2 \\
& > 2\beta_1(z_2 - \lambda) - 2\beta_2(z_2 - \lambda) + (z_2 - \lambda)^2 + (|z_1| - \lambda)^2 \\
& > 0,
\end{align*}
where the second to last inequality follows from the fact that $|z_1| - \lambda > z_2 - \lambda$, and the last inequality follows because $\beta_1 > \beta_2$.

Next, for any $\lambda \in [z_2, |z_1|)$,
\[
nL_2(\lambda) = (\beta_1 + (|z_1| - \lambda)_+)^2 + \beta_2^2 > \beta_1^2 + \beta_2^2.
\]
Thus, $\lambda^*_2 \geq \max_{1 \leq j \leq 2} |z_j|$.  From Theorem~\ref{l:s-sparse}, it follows that no deterioration occurs in this case.

\noindent \textbf{Case 3b: $\mathbf{z_1 < 0}$, $\mathbf{z_2 > 0}$, $\mathbf{|z_1| < |z_2|}$}

For any $\lambda \in [0, |z_1|)$,
\[
nL_2(\lambda) - nL_2(z_2) = (\beta_1 + (|z_1| - \lambda)_+)^2 + (\beta_2 - (z_2 - \lambda)_+)^2  - (\beta_1^2 + \beta_2^2) > 0,
\]
where the last inequality follows from a similar argument to that used in case 3a.

Next, for any $\lambda \in [|z_1|, z_2)$, 
\[
nL_2(\lambda) = \beta_1^2 + (\beta_2 - (z_2 - \lambda))^2.
\]
By computing the derivative, it follows that
\[
\argmin_{\lambda \in [|z_1|, z_2)} nL_2(\lambda) = \left\{
  \begin{array}{l l}
    z_2 - \beta_2 & \quad \text{if $0 \leq \beta_2 \leq z_2 - |z_1|$}\\
    |z_1| & \quad \text{if $z_2 - |z_1| < \beta_2$}
  \end{array} \right..\]
Comparing the loss for these values of $\lambda$ to the loss at $\lambda = z_2$, it follows that
\[ 
\lambda_2^* = \argmin_{\lambda \in [|z_1|, z_2)} nL_2(\lambda).
\]

\noindent \textbf{Case 3c: $\mathbf{z_1 > 0}$, $\mathbf{z_2 < 0}$, $\mathbf{|z_1| > |z_2|}$}

For any $\lambda \in [0, |z_2|)$,
\[
nL_2(\lambda) = (\beta_1 - (z_1 - \lambda))^2 + (\beta_2 + (z_2 - \lambda))^2
\]
and
\[
\frac{\partial nL_2(\lambda)}{\partial \lambda} =
  2((\beta_1 - \beta_2) - (z_1 + z_2) + 2 \lambda).
\]
Since this is an increasing function of $\lambda$, it follows that
\[
\argmin_{\lambda \in [0, |z_2|)} nL_2(\lambda) = \left\{
  \begin{array}{l l}
    |z_2| & \quad \text{if $0 \leq \beta_1 - \beta_2 \leq z_1 - |z_2|$}\\
    \frac{(z_1 + |z_2|) - (\beta_1 - \beta_2)}{2} & \quad \text{if $z_1 - |z_2| < \beta_1 - \beta_2 \leq z_1 + |z_2|$} \\
    0 & \quad \text{if $ z_1 + |z_2| < \beta_1 - \beta_2$}
  \end{array} \right..\]

Next, for any $\lambda \in (|z_2|, z_1)$,
\[
\frac{\partial nL_2(\lambda)}{\partial \lambda} =
  2(\beta_1 - z_1 + \lambda) \geq 0.
\]
Thus,
\[
\argmin_{\lambda \in [0, |z_2|)} nL_2(\lambda) = \left\{
  \begin{array}{l l}
    z_1 - \beta_1 & \quad \text{if $0 \leq \beta_1 \leq z_1 - |z_2|$}\\
    |z_2| & \quad \text{if $z_1 - |z_2| < \beta_1$} \\
  \end{array} \right..\]

Comparing the loss values at the local optima, it follows from a tedious but straightforward calculation that
\[
\lambda_2^* = \left\{
  \begin{array}{l l}
    z_1 - \beta_1 & \quad \text{if $0 \leq \beta_1 \leq z_1 - |z_2|$}\\
    |z_2| & \quad \text{if $z_1 - |z_2| < \beta_1$ and $0 < \beta_1 - \beta_2$} \\
    \frac{(z_1 + |z_2|) - (\beta_1 - \beta_2)}{2} & \quad \text{if $z_1 - |z_2| < \beta_1$ and $z_1 - |z_2| < \beta_1 - \beta_2 < z_1 + |z_2|$} \\
    0 & \quad \text{if $\beta_1 - \beta_2 > z_1 + |z_2|$} \\
  \end{array} \right..\]

\noindent \textbf{Case 3d: $\mathbf{z_1 > 0}$, $\mathbf{z_2 < 0}$, $\mathbf{|z_1| < |z_2|}$}

For any $\lambda \in [0, z_1)$,
\[
\frac{\partial nL_2(\lambda)}{\partial \lambda} =
  2((\beta_1 - \beta_2) - (z_1 + |z_2|) + 2\lambda) \geq 0
\]
This is an increasing function of $\lambda$.  Thus, the minimum occurs at $\lambda = 0$ if the derivative is positive at this point.  Otherwise, the minimum occurs at the point where the derivative is zero,
\[
\lambda = \frac{(z_1 + z_2) - (\beta_1 - \beta_2)}{2}.
\]
This implies that
\[
\argmin_{\lambda \in [0, z_1)} nL_2(\lambda) = \left\{
  \begin{array}{l l}
    \frac{z_1 + |z_2| - (\beta_1 - \beta_2)}{2} & \quad \text{if $0 \leq \beta_1 - \beta_2 \leq z_1 + |z_2|$}\\
    0 & \quad \text{if $z_1 + |z_2| \leq \beta_1 - \beta_2$} \\
  \end{array} \right..\]
Next, for any $\lambda \in (z_1, |z_2|)$,
\[
nL_2(\lambda) = \beta_1^2 + (\beta_2 + |z_2| - \lambda)^2 > \beta_1^2 + \beta_2^2 = nL_2(|z_2|).
\]
Thus, it is never optimal to choose $\lambda$ in this interval.  It follows that $\lambda_2^* = \argmin_{\lambda \in [0, z_1)} nL_2(\lambda)$ or $\lambda_2^* = |z_2|$.

\vspace{5mm}
\noindent \textbf{Probability of deterioration.}

We can numerically estimate the probability of deterioration
using simulations by computing $\lambda_2^*$ for each realization and determining whether or not the conditions of Theorem~\ref{T:prob-det-s-sparse}
are satisfied.  Table~\ref{tab:prob-det-s2} reports the estimated probability of deterioration based on 10,000 realizations with $\beta_1 = 3$, $\beta_2 = 1$ and varying values of $\sigma$ and $p$.  As was the case for $s = 1$, the results suggest that the probability of deterioration is close to one for a sufficiently high signal to noise ratio and large $p$ when $p_0 = 2$.

\begin{table}
\caption{\label{tab:prob-det-s2} The estimated probability of deterioration when $p_0 = 2$, $\beta_1 = 3$ and $\beta_2 = 1$
for varying $\sigma$ and $p$ computed over 10,000 realizations.}
\centering
\fbox{%
\begin{tabular}{l ccccc}
& \multicolumn{5}{c}{Probability of Deterioration}\\
$\sigma$  & $p=3$       & $p=5$       & $p=10$       & $p=50$       & $p=100$\\
0.5	&	0.4001	&	0.7423	&	0.9313	&	0.9959	&	0.9976	\\
1	&	0.3631	&	0.6906	&	0.8836	&	0.9645	&	0.9697	\\
3	&	0.2479	&	0.4965	&	0.6829	&	0.8431	&	0.8758	\\
9	&	0.1472	&	0.3224	&	0.4936	&	0.6890	&	0.7218	\\
\end{tabular}}
\end{table}

\section{Optimal APSE Ratio}\label{app:mse}
\begin{figure}[H]
\centering
\caption{Optimal APSE ratio for $p$ predictors compared to the 6 true predictors over 1000 realizations as a function of $p$.  The ``High SNR'' and ``Low SNR'' settings correspond to $\sigma^2=9$ and $\sigma^2=625$, respectively.}
\label{f:sim_mse}
\subfigure[High SNR]{
\resizebox{6.5cm}{!}{\includegraphics{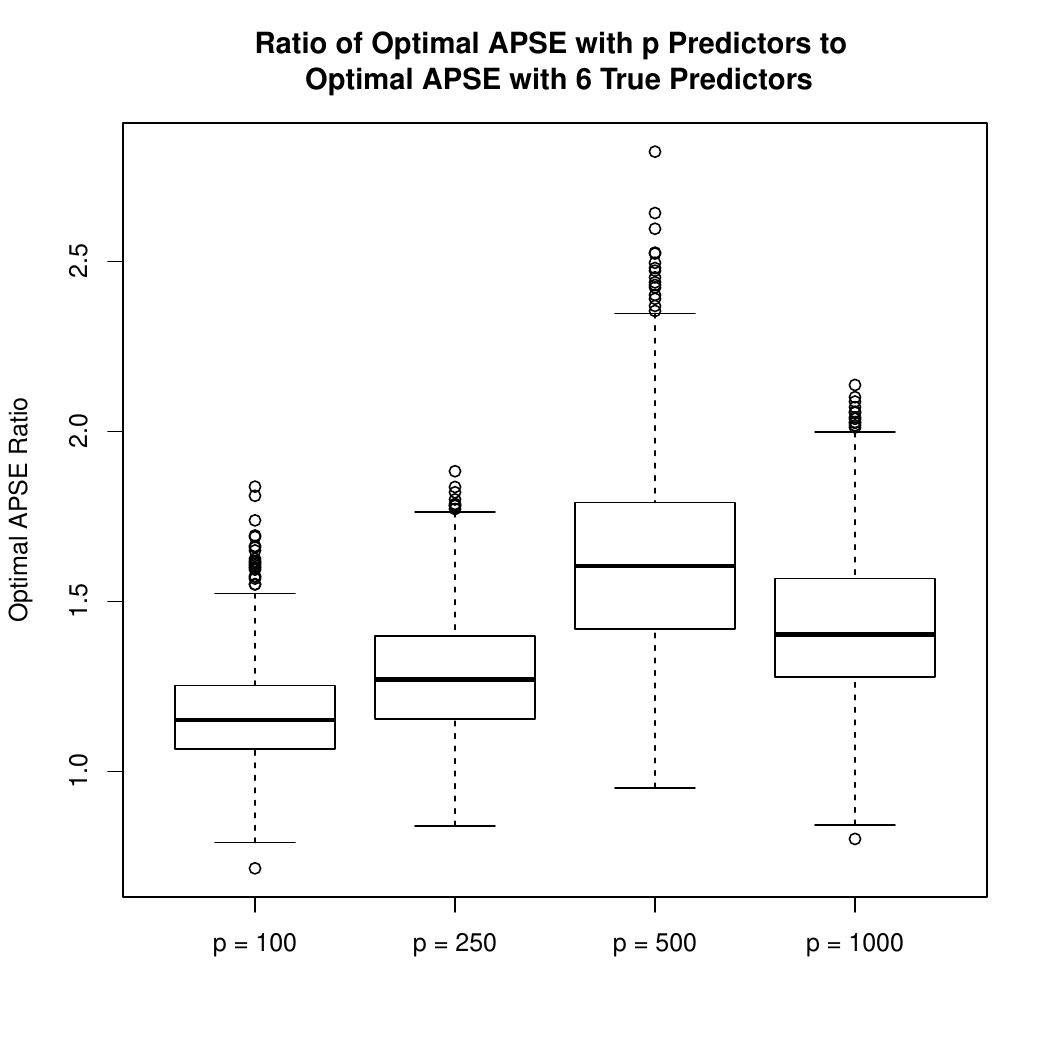}}}
\subfigure[Low SNR]{
\resizebox{6.5cm}{!}{\includegraphics{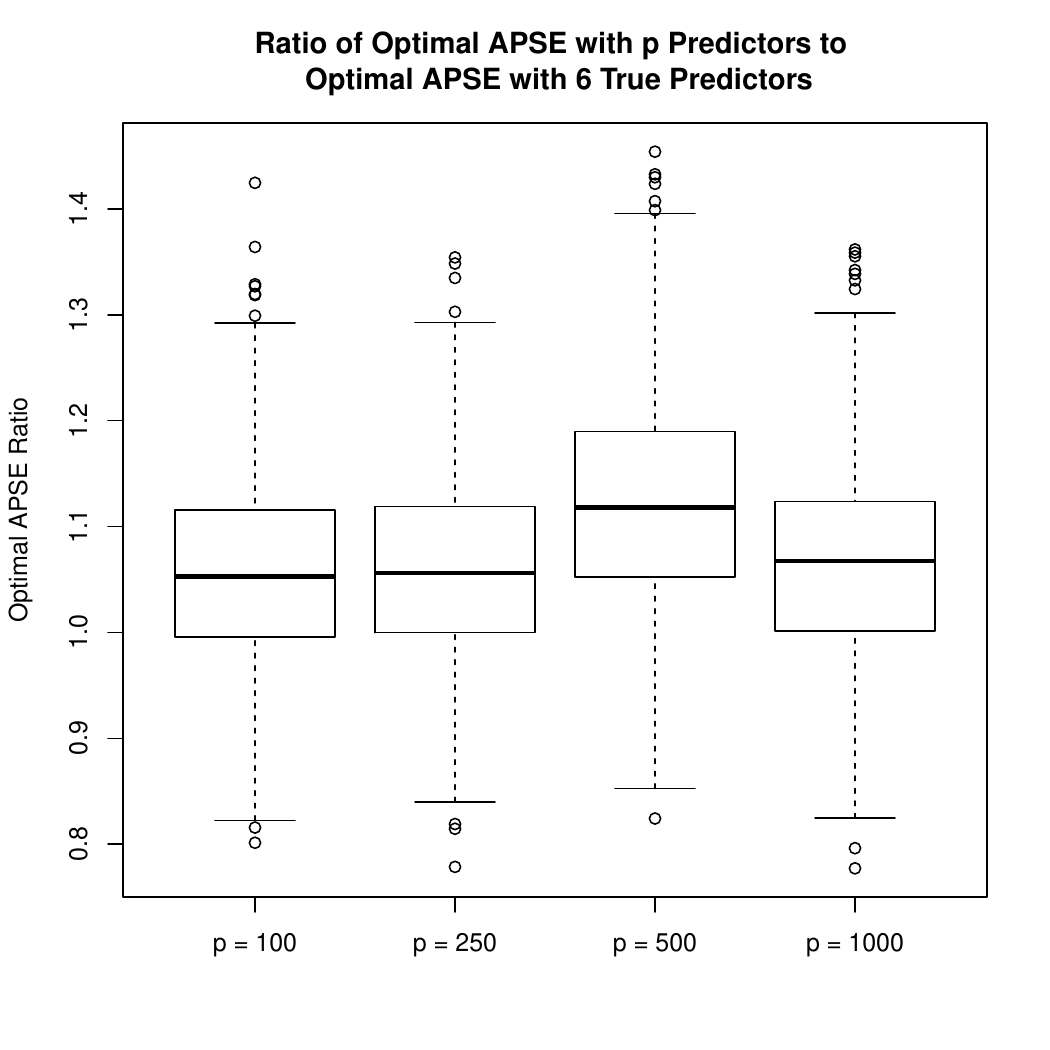}}}
\end{figure}

Here we return to the independent predictors example in section~\ref{S:indep-pred}.  To study the behavior of the optimal APSE, we evaluate the APSE for each realization on a simulated test set.  Figure~\ref{f:sim_mse} presents boxplots of the ratios of the estimated optimal APSE with $p$ predictors to the estimated optimal APSE with the six true predictors where $p$ is taken to be 100, 250, 500, and 1000 and $n = 100$.  A comparison of this figure to the median optimal loss ratios presented in Figure~\ref{f:ratios_pgn} demonstrates that
while deterioration is still observed, the optimal APSE ratios can be smaller than the optimal loss ratios.  To understand why this is the case, note that the APSE is equal to
\[
\frac{1}{n}||\vy^* - \hat \vy||^2 = \frac{1}{n} \Big( ||\vmu^* - \hat \vmu||^2 + 2 (\vmu^* - \hat \vmu)^T \varepsilon^* + ||\varepsilon^*||^2 \Big),
\]
where $\cdot^*$ is with respect to an independent test set.  Thus, the optimal APSE ratios can be smaller than the optimal loss ratios due to the presence of additional terms in both the numerator and denominator of the APSE ratio.

These figures also suggest that the deterioration pattern is less well-behaved when $p > n$ than it is when $p < n$, which is consistent with the results found in Section~\ref{S:indep-pred}.

\bibliography{myreferences2}
\bibliographystyle{imsart-nameyear}

\end{document}